\input ./style/arxiv-general.cfg
\documentclass[aos,MSNbibl,dvips]{arximspdf}
\makeatletter
   \@ifpackageloaded{graphicx}{}{\usepackage{graphicx}}
\makeatother
\usepackage{mathrsfs}
\usepackage{url,breakurl}

\doi{10.1214/14-AOS1308}
\volume{43}
\issue{3}
\pubyear{2015}
\firstpage{1243}
\lastpage{1272}
\docsubty{FLA}

\makeatletter

\newcommand{\rrvert}{\vert}

\newcommand{\llvert}{\vert}
\newcommand{\eqref}[1]{(\ref{#1})}

\theoremstyle{definition}

\newcommand{\argmin}{\mathop{\arg\min}}

\newproclaim{definition}{Definition}
\newtheorem{theorem}{Theorem}
\newtheorem{corollary}{Corollary}
\newtheorem{lemma}{Lemma}

\newtheorem{condition}{Condition}
\newtheorem{proposition}{Proposition}

\newcommand{\bdelta}{\bolds\delta}
\newcommand{\btheta}{\bolds\theta}
\newcommand{\bmu}{\bolds\mu}
\newcommand{\bX}{{\mathbf X}}
\newcommand{\be}{{\mathbf e}}
\newcommand{\bv}{{\mathbf v}}
\newcommand{\bA}{{\mathbf A}}
\newcommand{\by}{{\mathbf y}}

\newcommand{\bw}{\mathbf w}
\newcommand{\bB}{\mathbf B}
\newcommand{\bZ}{\mathbf Z}
\newcommand{\bzero}{\mathbf 0}
\newcommand{\bveps}{\bolds{\varepsilon}}
\newcommand{\bSig}{\bolds{\Sigma}}
\newcommand{\bOmega}{\bolds{\Omega}}

\newcommand{\var}{\operatorname{var}}
\newcommand{\cov}{\operatorname{cov}}
\newcommand{\supp}{\operatorname{supp}}

\makeatother

\begin{document}
\begin{frontmatter}

\title{Innovated interaction screening for high-dimensional nonlinear
classification\thanksref{T1}}
\runtitle{Innovated interaction screening}
\thankstext{T1}{Supported by NSF CAREER Award DMS-11-50318 and USC
Marshall Summer Research Funding.}

\begin{aug}
\author[A]{\fnms{Yingying}~\snm{Fan}\ead
[label=e1]{fanyingy@marshall.usc.edu}},
\author[B]{\fnms{Yinfei}~\snm{Kong}\corref{}\ead
[label=e2]{yinfeiko@usc.edu}},
\author[A]{\fnms{Daoji}~\snm{Li}\ead[label=e3]{daojili@marshall.usc.edu}}
\and
\author[C]{\fnms{Zemin}~\snm{Zheng}\ead[label=e4]{zeminzhe@usc.edu}}
\runauthor{Fan, Kong, Li and Zheng}
\affiliation{University of Southern California}
\address[A]{Y. Fan\\
D. Li\\
Data Sciences and Operations Department\\
Marshall School of Business\\
University of Southern California\\
Los Angeles, California 90089\\
USA\\
\printead{e1}\\
\phantom{E-mail:\ }\printead*{e3}}

\address[B]{Y. Kong\\
Department of Preventive Medicine\\
Keck School of Medicine\\
University of Southern California\\
Los Angeles, California 90089\\
USA\\
\printead{e2}}

\address[C]{Z. Zheng\\
Department of Mathematics\\
University of Southern California\\
Los Angeles, California 90089\\
USA\\
\printead{e4}}
\end{aug}

%
\received{\smonth{10} \syear{2014}}


\begin{abstract}
This paper is concerned with the problems of interaction screening and
nonlinear classification in a high-dimensional setting. We propose a
two-step procedure, IIS-SQDA, where in the first step an innovated
interaction screening (IIS) approach based on transforming the original
$p$-dimensional feature vector is proposed, and in the second step a
sparse quadratic discriminant analysis (SQDA) is proposed for further
selecting important interactions and main effects and simultaneously
conducting classification. Our IIS approach screens important
interactions by examining only $p$ features instead of all two-way
interactions of order $O(p^2)$. Our theory shows that the proposed
method enjoys sure screening property in interaction selection in the
high-dimensional setting of $p$ growing exponentially with the sample
size. In the selection and classification step, we establish a sparse
inequality on the estimated coefficient vector for QDA and prove that
the classification error of our procedure can be upper-bounded by the
oracle classification error plus some smaller order term. Extensive
simulation studies and real data analysis show that our proposal
compares favorably with existing methods in interaction selection and
high-dimensional classification.
\end{abstract}

%

\end{frontmatter}

\section{Introduction}\label{sec:Intro}

Classification, aiming at identifying to which of a set of categories a
new observation belongs, has been frequently encountered in various
fields such as genomics, proteomics, face recognition, brain images,
medicine and machine learning. In recent years, there has been a
significant surge of interest in interaction selection in
classification due to the importance of interactions in statistical
inference and contemporary scientific discoveries. For instance, in
genome-wide association studies, it has been increasingly recognized
that gene--gene interactions and gene--environment interactions
substantially influence the risk of developing a human disease \cite
{kooperberg2009structures}. Ignoring these interactions could
potentially lead to misunderstanding about disease mechanisms as they
are potential sources of the missing heritability \cite{pan2011adaptive}.

Identification of interactions is challenging even when the number of
predictors~$p$ is moderately large compared to the sample size $n$, as
the number of all possible pairwise interaction effects is of order
$O(p^2)$. This problem becomes even more challenging in the
high-dimensional setting where $p$ can be much larger than $n$. It is
well known that the classical low-dimensional classification method
cannot be directly used for high-dimensional classification for at
least three reasons. First, many popular classifiers, such as linear
discriminant analysis (LDA) and quadratic discriminant analysis (QDA),
are inapplicable when $p$ exceeds $n$ because of the singularities of
the sample covariance matrices. Second, when $p$ is large, it is
commonly believed that only a subset of the $p$ features contribute to
classification. Classification using all potential features may cause
difficulty in interpretation and degrade the classification performance
due to the noise accumulation in estimating a large number of
parameters \cite{fan2008high}. Third, the computational cost may be
extremely high when the dimensionality is ultra-high. For example, with
$p=1000$ features, the dimensionality is about half million if all
possible pairwise interactions are included in classification.

In recent years, significant efforts have been made to develop
effective high-dimensional classification methods. The most commonly
imposed assumption is sparsity, leading to sparse classifiers.
Tibshirani et al. \cite{tibshirani2002diagnosis} introduced the nearest
shrunken centroids classifier, and Fan and Fan \cite{fan2008high}
proposed features annealed independent rules, both of which ignore
correlations among features to reduce the dimensionality of parameters.
Shao et al. \cite{shao2011sparse} proposed and studied a sparse LDA
method, which directly plugs the sparse estimates of the
covariance matrix and mean vector into the linear classifier. Cai and
Liu \cite{cai2011direct} introduced a direct approach to sparse LDA by
estimating the product of the precision matrix and the mean difference
vector of two classes, through constrained $L_1$ minimization. In an
independent work, Mai
et al. \cite{mai2012direct} also proposed a direct approach to sparse
LDA, called DSDA, by reformulating the LDA problem as a penalized least
squares regression.
Fan et al. \cite{fan2013optimal} considered HCT classifier
for high-dimensional Gaussian classification with sparse precision
matrix when the signals are rare and weak, and studied its optimality.
A commonality of these aforementioned methods is that the underlying
true classifier is assumed to be linear, and thus they belong to the
class of sparse LDA methods.

A key assumption for LDA is that observations from different classes
share the same correlation structure.
Although this assumption can significantly reduce the number of
parameters need to be estimated, it can be easily violated in real
applications. In addition, linear classifiers are not capable of
identifying important interaction effects between features and thus can
lead to inferior
feature selection and classification results, and consequently,
misleading interpretations when the classification boundary is nonlinear.
For instance, in a two-class Gaussian classification problem, when two
classes have equal mean vectors but different covariance matrices,
linear classifiers can perform no better than random guessing.

To gain some insight into the importance of interactions in
classification, let us look at a simple example. Consider a two-class
Gaussian classification problem
with the Bayes rule
%
\begin{eqnarray}
\label{eq:Qfun-example} Q({\mathbf z}) &=& - 0.3Z_{10}^2-0.15Z_{10}Z_{30}
-0.15Z_{10}Z_{50} - 0.3Z_{30}^2
\nonumber
\\[-8pt]
\\[-8pt]
\nonumber
&&{}-0.15Z_{30}Z_{50} - 0.3Z_{50}^2 +
1.74913,
\end{eqnarray}
which classifies a new observation ${\mathbf z}$ to class 1 if and only if
$Q({\mathbf z})>0$. Thus there are no main effects, and there are three
variables, $Z_{10}, Z_{30}$ and $Z_{50}$, contributing to interactions.
We simulated data in the same way as model 2 in Section~\ref{sec:
Study-estimated-Omega}, except that the mean vector in each class is
zero. See Section~\ref{sec: Study-estimated-Omega} for more details.
Table~\ref{table: motivation-example} lists the performance of
different classification methods, including penalized logistic
regression (PLR), DSDA, our proposal (IIS-SQDA) and the oracle
procedure (Oracle). The oracle procedure uses the information of the
true underlying sparse model and thus is a low-dimensional QDA. As
expected, both linear classifiers, PLR and DSDA, perform no better than
random guessing. Table~\ref{table: motivation-example} also shows the
variable selection results for main effects and interactions, with
FP.main standing for false positives of main effects, and FP.inter and
FN.inter standing for false positives and false negatives of
interaction effects, respectively. It is seen that with appropriate
selection of interaction effects, the classification performance can be
improved significantly.

\begin{table}
\caption{The means and standard errors (in parentheses) of various
performance measures for different classification methods over 100
replications where the Bayes rule is given in \protect\eqref{eq:Qfun-example}.
Sample size in each class is $100$, and the number of features $p$ is
$200$}\label{table: motivation-example}
\begin{tabular*}{\textwidth}{@{\extracolsep{\fill}}lcccc@{}}
\hline
\textbf{Measure} & \textbf{PLR} & \textbf{DSDA} & \textbf{IIS-SQDA} & \textbf{Oracle} \\
\hline
MR (\%) & 49.95 (0.05) & 49.87 (0.09) & 26.03 (0.31) & 23.99 (0.08) \\
FP.main & 49.66 (4.93) & 74.29 (6.45) & 3.16 (0.82) & 0 (0) \\
FP.inter & -- & -- & 0.55 (0.14) & 0 (0) \\
FN.inter & -- & -- & 0.15 (0.05) & 0 (0) \\
\hline
\end{tabular*}
\end{table}

In this paper we consider two-class classification with possibly
unequal covariance matrices. Under some sparsity assumption on the main
effects and interactions, we propose a two-stage classification
procedure, where we first reduce the number of interactions to a
moderate order by a new interaction screening approach, and then
identify both important main effects and interactions using some
variable selection techniques. Our interaction screening approach is
motivated by a result, which will be formally demonstrated in our
paper, that if an interaction term, say $Z_1Z_2$, appears in Bayes
decision rule, then after appropriately transforming the original
features, the resulting new feature $\widetilde{Z}_1$ (and $\widetilde
{Z}_2$) has different variances across classes. Thus the original
problem of screening $O(p^2)$ pairwise interaction effects can be
recast as the problem of comparing variances of only $p$ variables,
which can be solved by some variance test procedures such as the
$F$-test or the SIRI method proposed in \cite{jiang2014sliced}.\vadjust{\goodbreak} The
similar idea of interaction screening has also been considered in \cite
{jiang2014sliced} under the model setting of sliced inverse index
model. Hereafter, we refer to $Z_i$ as an interaction variable if an
interaction term involving $Z_i$ appears in Bayes rule. After obtaining
interaction variables in the first step, we reconstruct interaction
terms based on these screened interaction variables, and then use
recent advances in variable selection literature to further select
important ones from the pool of all main effects and reconstructed
interactions. Under some mild conditions, we prove that with
overwhelming probability, all active interaction variables will be
retained using our screening procedure. For the second step of
selection and classification, we first establish a sparse inequality
\cite{van2008high}, which shows the consistency of the estimated
coefficient vector of QDA, then further prove that the classification
error of IIS-SQDA is upper-bounded by the oracle classification error
plus a smaller order term. Our numerical studies demonstrate the fine
performance of the proposed method for interaction screening and
high-dimensional classification.

The main contributions of this paper are as follows. First, we
introduce an interaction screening approach, which has been proved to
enjoy sure screening property. Second, our classification method does
not rely on the linearity assumption, which makes our method more
applicable in real applications.
Third, our proposed classification procedure is adaptive in the sense
that it automatically chooses between sparse LDA and sparse QDA. If the
index set of screened interaction variables is empty in the first step,
or if the index set in the first step is nonempty but none of the
interaction terms is selected in the second step, then sparse LDA will
be used for classification; otherwise, sparse QDA will be used for
classification. Fourth, we provide theoretical justifications on the
effectiveness of the proposed procedure. 

The remaining part of the paper will unfold as follows. Section~\ref
{sec: Model} introduces the model setting and motivation. Section~\ref
{sec: Screening} proposes the innovated interaction screening approach
and studies its theoretical property. Section~\ref{sec: Selection}
considers post-screening variable selection. Section~\ref{sec:
Numerical} presents the results of extensive simulation studies and a
real data example. Section~\ref{sec: Discussion} concludes with some
discussion.
Section~\ref{sec: SecA} collects all proofs for the main theorems.
Additional proofs are provided in the supplementary material \cite
{fan2014Supp}.

\section{Model setting and motivation} \label{sec: Model}

Our interaction screening approach is motivated from the problem of
two-class Gaussian classification, where the $p$-dimensional feature
vector ${\mathbf z}=(Z_1, \ldots, Z_p)^T$ follows a mixture distribution
%
\begin{equation}
\label{model} {\mathbf z}= \Delta{\mathbf z}^{(1)} + (1-\Delta){\mathbf
z}^{(2)}
\end{equation}
with ${\mathbf z}^{(k)}$ a Gaussian random vector with mean $\bmu_k$ and
covariance matrix $\bSig_k$, $k=1,2$, and the class label $\Delta$
following a Bernoulli distribution with probability of success $\pi$.
Without loss of generality, assume that $\bmu_2=0$.
Under this model setting, the Bayes rule admits the following form:
%
\begin{equation}
\label{eq:Qfun-Bayes} Q({\mathbf z})=\tfrac{1}{2}{\mathbf z}^T\bOmega{
\mathbf z}+ \bdelta ^T {\mathbf z}+ \zeta,
\end{equation}
where $\bOmega=\bSig_2^{-1}-\bSig_1^{-1}$, $\bdelta=\bSig
_1^{-1}\bmu_1$
and $\zeta$ is some constant depending only on $\pi$, $\bmu_1$ and
$\bSig_k$, $k=1, 2$.
A new observation ${\mathbf z}$ is classified into class $1$ if and
only if $
Q({\mathbf z})>0$.

When covariance matrices $\bSig_1$ and $\bSig_2$ are the same, the
above Bayes rule takes the linear form $Q({\mathbf z}) = \bdelta^T
{\mathbf z}+ \zeta$,
which is frequently referred to as the Fisher's LDA and belongs to the
family of linear classifiers. As discussed in the \hyperref[sec:Intro]{Introduction}, linear
classifiers may be inefficient or even fail when the true
classification boundary is nonlinear. 
Moreover, linear classifiers are incapable of selecting important
interaction terms when the covariance matrices are different across two
classes. For the ease of presentation, hereafter we mean interaction in
the broad sense of the term, not just the two-way interactions
$Z_jZ_{\ell}$ with $j\neq\ell$, but also the quadratic terms $Z_j^2$.
So there are $p(p+1)/2$ possible interactions in total under our
definition. Throughout this paper we call $Z_jZ_{\ell}$, $1\leq j, \ell
\leq p$ an active interaction if its coefficient is nonzero in (\ref
{eq:Qfun-Bayes}), and we call $Z_j$ an interaction variable if there
exists some $\ell\in\{1,2,\ldots, p\}$ such that $Z_jZ_{\ell}$ is an
active interaction. Selecting important ones from the large number of
interactions is interesting yet challenging. We next discuss our
proposal for interaction screening. 

From \eqref{eq:Qfun-Bayes}, one can observe that an interaction term
$Z_jZ_{\ell}$ is an active interaction if and only if $\bOmega_{j\ell
}\neq0$. Here we use $\bA_{j\ell}$ to denote the $(j,\ell)$ element of
any matrix~$\bA$. This observation motivates us to select active
interactions by recovering the support of $\bOmega$. Denote the index
set of interaction variables by
%
\begin{equation}
\label{eq: setI} \mathcal{I}=\{1\leq j\leq p\dvtx Z_jZ_{\ell}
\mbox{ is an active interaction for some } 1\leq\ell\leq p\}.
\end{equation}
In light of \eqref{eq:Qfun-Bayes}, the above set can also be written as
$\mathcal{I}=\{1\leq j\leq p\dvtx \bOmega_{j\ell}\neq0 \mbox{ for
some } 1\leq\ell\leq p\}$. If the index set $\mathcal{I}$ can be
recovered, then all active interactions can be reconstructed. For this
reason, we aim at developing an effective method for screening the
index set $\mathcal{I}$.

In a high-dimensional setting, to ensure the model identifiability and
to enhance the model fitting accuracy and interpretability, it is
commonly assumed that only a small number of interactions contribute to
classification. Thus we impose the sparsity assumption that there are
only a small number of active interactions. Equivalently, we can assume
that $\bOmega$ is highly sparse with only $q = o(\min\{n,p\})$ rows
(and columns, by symmetry) being nonzero, where $n$ is the total sample
size. Denote $\bSig_k^{-1}$ by $\bOmega_k$, $k=1, 2$. Without loss of
generality, we write $\bOmega$ as
%
\begin{eqnarray}
\label{B} \bOmega=\bOmega_2 - \bOmega_1 =\pmatrix{ \bB& {\mathbf0}
\vspace*{2pt}\cr
{\mathbf0}^T& {\mathbf0} },
\end{eqnarray}
where $\bB$ is a $q\times q$ symmetric matrix with at least one nonzero
element in each row. We remark that the block structure in (\ref{B}) is
just for the simplicity of presentation, and we do not require that the
locations of nonzero rows of $\bOmega$ are known. In fact, we will
develop a method to estimate the indices of these nonzero rows.
Note that the set $\mathcal{I}$ can be further written as
\[
\mathcal{I}=\{1\leq j\leq q\dvtx \bB_{j\ell}\neq0 \mbox{ for some } 1\leq
\ell\leq q\}. 
\]
Thus interaction screening is equivalent to finding the indices of
features related to $\bB$.

Identifying the index set $\mathcal{I}$ is challenging when $p$ is
large. We overcome this difficulty by decomposing $\mathcal{I}$ into
two subsets. Let
\[
\mathcal{I}_1= \{j \in\mathcal{I} \mbox{ and } \bB_{jj}
\leq0\}, \qquad\mathcal{I}_2= \{j \in\mathcal{I} \mbox{ and }
\bB_{jj}> 0\}.
\]
Then $\mathcal{I}=\mathcal{I}_1\cup\mathcal{I}_2$. This allows us to
estimate $\mathcal{I}$ by dealing with $\mathcal{I}_1$ and $\mathcal
{I}_2$ separately.

First consider $\mathcal{I}_1$. Our main idea is to use the
transformation $\tilde{{\mathbf z}} = \bOmega_1 {\mathbf z}$. Denote
by $\tilde{{\mathbf z}
}^{(k)} = \bOmega_1 {\mathbf z}^{(k)}$ the transformed feature vector from
class $k$ with $k=1,2$. Then $\cov(\tilde{{\mathbf z}}^{(1)})=\bOmega
_1$ and
$\cov(\tilde{{\mathbf z}}^{(2)})= \bOmega_1\bSig_2\bOmega_1$.
It follows from linear algebra that the difference of the above two
covariance matrices takes the following form:
%
\begin{equation}
\label{ezero} \widetilde{\bSig}_1\equiv\bOmega_1
\bSig_2\bOmega_1 - \bOmega_1 = \bOmega
\bSig_2\bOmega- \bOmega =\pmatrix{
\bB\bSig_{2}^{(11)}\bB-\bB& {\mathbf0}
\vspace*{2pt}\cr
{\mathbf0}^T& {\mathbf0} },
\end{equation}
where $\bSig_2^{(11)}$ is the $q\times q$ 
principal submatrix of $\bSig_2$ corresponding to matrix $\bB$.
We will show that if $j\in\mathcal{I}_1$, then the $j$th entry in transformed
feature vector $\tilde{{\mathbf z}}$ has different variances across two
classes. To this end,
let $\be_j$ be a unit vector with $j$th component 1 and all other
components 0. Then it follows from the positive definiteness of $\bSig
_2^{(11)}$ that $(\bB\be_j)^T\bSig_2^{(11)}(\bB\be_j)$ is positive
for any $j \in\mathcal{I}_1$. Since $\bB_{jj}\leq0$ for any $j\in
\mathcal{I}_1$, the $j$th diagonal element of $\widetilde{\bSig}_1$ is
positive by noting that
%
\begin{equation}
\label{eq:vardiff1} (\widetilde{\bSig}_1)_{jj}=(\bB
\be_j)^T\bSig_2^{(11)}(\bB\be
_j)-\bB_{jj}.
\end{equation}
This gives a set inclusion
%
\begin{equation}
\label{eq: set-inclusion-1} \mathcal{I}_1 \subset\mathcal{A}_1
\equiv\bigl\{j\dvtx (\widetilde{\bSig }_1)_{jj} \neq0\bigr
\}.
\end{equation}
Observing that $(\widetilde{\bSig}_1)_{jj} $ is the difference of
between-class variances of the $j$th transformed variable, that is,
%
\begin{equation}
\label{eq:vardiff2} (\widetilde{\bSig}_1)_{jj} = \var\bigl(
\be_j^T\tilde{\mathbf z}^{(2)}\bigr) - \var\bigl(
\be _j^T\tilde{\mathbf z}^{(1)}\bigr),
\end{equation}
the index set $\mathcal{A}_1$ can be obtained by examining which
features have different variances across two classes after the transformation.

We further remark that the variance difference between $\be_j^T\tilde
{\mathbf z}^{(2)}$ and $\be_j^T\tilde{\mathbf z}^{(1)}$ records the
accumulated
contributions of the $j$th feature to the interaction. To understand
this, note that if $\bSig_2^{(11)}$ has the smallest eigenvalue bounded
from below by a positive constant $\tau_1$, then \eqref{eq:vardiff1}
and \eqref{eq:vardiff2} together ensure that
\[
\var\bigl(\be_j^T\tilde{\mathbf z}^{(2)}\bigr)
- \var\bigl(\be_j^T\tilde {\mathbf z}^{(1)}
\bigr) \geq\tau _1\|\bB\be_j\|_2^2
- \bB_{jj}, 
\]
where $\|\cdot\|_2$ denotes the $L_2$ norm of a vector. In view of
\eqref{eq:Qfun-Bayes} and \eqref{B}, the $j$th column (and row, by
symmetry) of $\bB$ records all contributions of the $j$th feature to
interactions. Thus the more important the $j$th feature is to
interaction, the larger the variance difference.

Similarly, consider the transformation $\check{{\mathbf z}}= \bOmega
_2 {\mathbf z}$,
and define the matrix $\widetilde{\bSig}_2 = \bOmega_2 - \bOmega
_2\bSig
_1\bOmega_2$. Then $\widetilde{\bSig}_2$ is the difference between the
covariance matrices of transformed feature vectors $\check{{\mathbf
z}}^{(1)} =
\bOmega_2 {\mathbf z}^{(1)}$ and $\check{{\mathbf z}}^{(2)} = \bOmega
_2 {\mathbf z}^{(2)}$.
Using arguments similar to those in \eqref{eq: set-inclusion-1}, we get
another set inclusion
%
\begin{equation}
\label{eq: set-inclusion-2} \mathcal{I}_2 \subset\mathcal{A}_2 \equiv
\bigl\{j\dvtx (\widetilde{\bSig }_2)_{jj} \neq0\bigr\}.
\end{equation}
Similarly, the set $\mathcal{A}_2$ can be obtained by examining which
features have different variances across two classes after the
transformation based on $\bOmega_2$.

Combining \eqref{eq: set-inclusion-1} and \eqref{eq: set-inclusion-2}
leads to
%
\begin{equation}
\label{eq: set-inclusion-3} \mathcal{I}\subset\mathcal{A}_1\cup
\mathcal{A}_2.
\end{equation}
Meanwhile, by (\ref{ezero}) we have $\mathcal{A}_1 \subset\mathcal
{I}$. Similarly, we obtain $\mathcal{A}_2 \subset\mathcal{I}$.
Combining these results with the set inclusion (\ref{eq:
set-inclusion-3}) ensures that
%
\begin{equation}
\label{eq:set-inclusion-4} \mathcal{I} = \mathcal{A}_1\cup\mathcal{A}_2.
\end{equation}
This motivates us to find interaction variables by testing variances of
the transformed feature vectors $\widetilde{{\mathbf z}}$ and $\check
{{\mathbf z}}$
across two classes. Since the transformation based on precision matrix
$\bOmega_k$ is called innovation in the time series literature, we name
our method the \textit{innovated interaction screening} (IIS). 

The innovated transform has also been explored in other papers. For
example, Hall and Jin \cite{hall2010innovated} proposed the innovated
higher criticism based on the innovated transform on the original
feature vector to detect sparse and weak signals when the noise
variables are correlated,
and established an upper bound to the detection boundary. In two-class
Gaussian linear classification setting, Fan et al. \cite
{fan2013optimal} discussed in detail the advantage of innovated
transform. 
They showed that the innovated transform is best at boosting the
signal-to-noise ratio in their model setting. Detailed discussions
about innovated transform in the multiple testing context can be found
in \cite{jin2012comment}.

\section{Sampling property of innovated interaction screening}\label
{sec: Screening}

\subsection{Technical assumptions}\label{subsec: assumptions}

We study the sampling properties of IIS procedure in this section. In
our theoretical development, the Gaussian distribution assumption in
Section~\ref{sec: Model} will be relaxed to sub-Gaussian, but
implicitly, we still assume that our target classifier takes the form
\eqref{eq:Qfun-Bayes}. A random vector $\bw=(W_1, \ldots, W_p)^T\in
\mathbb{R}^p$ is sub-Gaussian if there exist some positive constants
$a$ and $b$ such that $P(|\bv^T\bw|>t) \leq a\exp(-bt^2)$
for any $t>0$ and any vector $\bv\in\mathbb{R}^p$ satisfying $\|\bv
\|
_2=1$. 
The following conditions will be needed for our theoretical development:

\begin{condition} [(Sub-Gaussian)] \label{con: distribution}
Both ${\mathbf z}^{(1)}$ and ${\mathbf z}^{(2)}$ are sub-Gaussian.
\end{condition}

\begin{condition} [(Bounds of eigenvalues)] \label{coneig}
There exists some positive constant $\tau_1$ and some positive sequence
$\tau_{2,p}$ depending only on $p$ such that the eigenvalues of $\bSig
_1$ and $\bSig_2$ satisfy
\[
\tau_1 \leq\lambda_{\min}(\bSig_k) \leq
\lambda_{\max}(\bSig_k) \leq \tau_{2,p}\qquad \mbox{for
} k = 1,2,
\]
where $\lambda_{\min}(\cdot)$ and $\lambda_{\max}(\cdot)$ denote the
smallest and largest eigenvalues of a matrix, respectively.
\end{condition}

\begin{condition} [(Distinguishability)] \label{condistin}
Denote by $\sigma^2_{j}, (\sigma_{j}^{(1)})^2$ and $(\sigma
_{j}^{(2)})^2$ the population variances of the $j$th covariates in
$\tilde{{\mathbf z}}$, $\tilde{{\mathbf z}}^{(1)}$ and $\tilde
{{\mathbf z}}^{(2)}$,
respectively. There exist some positive constants $\kappa$ and $c$ such
that for any $j \in\mathcal{A}_1$ with $\mathcal{A}_1$ defined in~\eqref{eq: set-inclusion-1}, it holds that
%
\begin{equation}
\label{eq: variance-ratio} \frac{\sigma^2_{j}}{(\sigma_{j}^{(1)})^{2\pi} (\sigma_{j}^{(2)})^{2(1
- \pi)}} \geq\exp \bigl(3c n^{-\kappa} \bigr).
\end{equation}
%
Moreover, the same inequality also holds for the $j$th covariates in
$\check{{\mathbf z}}$, $\check{{\mathbf z}}^{(1)}$ and $\check
{{\mathbf z}}^{(2)}$ when $j \in
\mathcal{A}_2$ with $\mathcal{A}_2$ defined in \eqref{eq: set-inclusion-2}.
\end{condition}

\begin{condition} [($K_p$-sparsity)] \label{conv}
For each $k=1,2$, the precision matrix $\bOmega_{k}$ is $K_p$-sparse,
where a matrix is said to be $K_p$-sparse if each of its row has at
most $K_p$ nonzero components with $K_p$ a positive integer depending
only on $p$. Moreover, $\|\bOmega_{k}\|_{\max}$ is bounded from above
by some positive constant independent of $p$, where $\|\cdot\|_{\max}$
is the elementwise infinity norm of a matrix.
\end{condition}

Condition \ref{con: distribution} is used to control the tail behavior
of the covariates.
Gaussian distribution and distributions with bounded support are two
special examples of sub-Gaussian distribution.

Condition \ref{coneig} imposes conditions on the eigenvalues of the
population covariance matrices $\bSig_1$ and $\bSig_2$. The lower bound
$\tau_1$ is a constant while the upper bound can slowly diverge to
infinity with $p$. So the condition numbers of $\bSig_1$ and $\bSig_2$
can diverge with $p$ as well. We remark that we need a constant lower
bound $\tau$ to exclude the case of perfect or nearly perfect
collinearity of features at the population level. On the technical
side, the constant lower bound $\tau_1$ ensures that $\tilde{\mathbf z}^{(k)}$
and $\check{{\mathbf z}}^{(k)}$ are still sub-Gaussian after transformation.

Condition \ref{condistin} is a signal strength condition which assumes
that for any $j \in\mathcal{A}_1$, the population variances of the
$j$th transformed feature $\widetilde{Z}_j$ are different enough across
classes, by noting that
%
\begin{equation}
\label{eq: Dj} D_j \equiv\log\sigma^2_j -
\sum_{k = 1}^2 \pi_k \log \bigl[
\bigl(\sigma_j^{(k)}\bigr)^2 \bigr] \geq 3c
n^{-\kappa}
\end{equation}
with $\pi_1=\pi$ and $\pi_2=1-\pi$ when $j \in\mathcal{A}_1$.
Meanwhile, it is clear from the definition of $\mathcal{A}_1$ that
$D_j$ is exactly $0$ when $j \in\mathcal{A}_1^c$ since the population
variances of the $j$th transformed covariate $\widetilde Z_j$ are the
same across classes. The same results hold for any feature with index
in $\mathcal{A}_2$, after transforming the data using $\bOmega_2$,
based on the second part of this condition.

Condition \ref{conv} is on the sparsity of the precision matrices,
which is needed for ensuring the estimation accuracy of precision
matrices. The same family of precision matrices has also been
considered in \cite{fan2013optimal} for high-dimensional linear
classification. 
Condition \ref{conv} also imposes a uniform upper bound for all
components of $\bOmega_{k}$. We note that we use this assumption merely
to simplify the proof, and our main results will still hold with a
slightly more complicated form when the upper bound diverges slowly %
with the number of predictors $p$. 

\subsection{Oracle-assisted IIS} \label{sec: Screening-true}

In this subsection, we consider IIS with known precision matrices,
which we call the oracle-assisted IIS. The case of unknown precision
matrices will be studied in the next subsection. The results developed
here are mainly of theoretical interests and will serve as a benchmark
for the performance of IIS with unknown precision matrices.

As introduced in Section~\ref{sec: Model}, IIS works with the
transformed feature vectors $\tilde{\mathbf z}= \bOmega_1{\mathbf z}$
and $\check{\mathbf z}
= \bOmega_2{\mathbf z}$ identically. For the ease of presentation we only
discuss in detail IIS based on the transformation $\tilde{{\mathbf z}} =
(\widetilde{Z}_1,\ldots, \widetilde{Z}_p)^T= \bOmega_1{\mathbf z}$.

Suppose we observe $n$ data points $\{({\mathbf z}_i^T, \Delta_i),
i=1, \ldots,
n\}$, $n_k$ of which are from class $k$ for $k=1, 2$. Write $\widetilde
{\bZ}=\bZ{\bOmega}_1$ as the transformed data matrix, where $\bZ
=({\mathbf z}
_1, \ldots, {\mathbf z}_n)^T$ is the original data matrix. To test
whether the
$j$th transformed feature $\widetilde{Z}_j$ has different variances
across two classes, we propose to use the following test statistic
introduced in \cite{jiang2014sliced}:
%
\begin{equation}
\label{eq: Dj-tilde} \widetilde{D}_j = \log\tilde{\sigma}^2_j
- \sum_{k = 1}^2 (n_{k}/n) \log
\bigl[\bigl(\tilde {\sigma}_j^{(k)}\bigr)^2
\bigr],
\end{equation}
where $\tilde{\sigma}_j^2$ denotes the pooled sample variance estimate
for $\widetilde{Z}_j$,
and $(\tilde{\sigma}_j^{(k)})^2$ is the within-class sample variance
estimate for $\widetilde{Z}_j$ in class $k$.
As can be seen from \eqref{eq: Dj-tilde}, $\widetilde{D}_j$ is expected
to be nonzero if variances of $\widetilde{Z}_j$
are different across classes. This test statistic was originally
introduced in \cite{jiang2014sliced} in the sliced inverse index model
setting for detecting important variables with pairwise or higher-order
interactions among $p$ predictors. The aforementioned paper recommends
the use of $\widetilde{D}_j$ in the initial screening step of their
proposed procedure, and proves the sure screening property of it under
some regularity conditions. 

Denote by
$\tilde{\tau}_{1,p} = \min\{\pi\tau_{2,p}^{-1} + (1 - \pi)\tau_1
\tau
_{2,p}^{-2}, 1\}$ and $\tilde{\tau}_{2,p} = \max\{\pi\tau_1^{-1} +
(1 -
\pi)\tau_1^{-2}\tau_{2,p} + \pi(1 - \pi)\tau_{1}^{-2} \|\bmu_1\|
_2^2,\exp(1)\}$.
The following proposition shows that the oracle-assisted IIS enjoys the
sure screening property in interaction selection under our model setting.

\begin{proposition}\label{Prop1}
Assume that Conditions \ref{con: distribution}--\ref{condistin} hold.
If $\log p = O(n^{\gamma})$ with $\gamma> 0$ and $\gamma+ 2\kappa<
1$, and $\tilde{\tau}_{1,p}^{-2} + \log^2(\tilde{\tau}_{2,p}) = o(n^{1
- 2\kappa- \gamma})$, then with probability at least $1 - \exp\{-C
n^{1 - 2\kappa}/[\tilde{\tau}_{1,p}^{-2} + \log^2(\tilde{\tau
}_{2,p})]\}
$ for some positive constant $C$, it holds that
\[
\min_{j \in\mathcal{A}_1} \widetilde{D}_j
\geq2c n^{-\kappa} \quad\mbox{and}\quad 
\max_{j \in\mathcal{A}_1^c}
\widetilde{D}_j \leq c n^{-\kappa},
\]
for large enough $n$, where $c$ is defined in Condition \ref
{condistin}. The same results also hold for the sets $\mathcal{A}_2$
and $\mathcal{A}_2^c$ with the test
statistics being calculated using data transformed by $\bOmega_2$.
\end{proposition}

The assumption $\tilde{\tau}_{1,p}^{-2} + \log^2(\tilde{\tau
}_{2,p}) =
o(n^{1 - 2\kappa- \gamma})$ restricts how fast the upper bound $\tau
_{2,p}$ in Condition \ref{coneig} can diverge with 
the number of predictors $p$.
Proposition~\ref{Prop1} entails that the oracle-assisted IIS can
identify all indices in $\mathcal{A}_1\cup\mathcal{A}_2$ with
overwhelming probability, by thresholding the test statistics
$\widetilde D_j$ with threshold chosen in the interval $(cn^{-\kappa},
2cn^{-\kappa})$. 
In view of \eqref{eq:set-inclusion-4}, Proposition~\ref{Prop1} gives
the variable selection consistency of the oracle-assisted IIS; that is,
the set of true interaction variables $\mathcal{I}$ can be selected
with asymptotic probability one. This result holds for ultra-high
dimensional $p$ satisfying $\log p = O(n^{\gamma})$ with $0 < \gamma<
1 - 2\kappa$. The key step in proving the theorem is to analyze the
deviation bound of $\widetilde{D}_j$ from its population counterpart
$D_j$. 
More details can be found in the supplementary material \cite{fan2014Supp}.

\subsection{IIS with unknown precision matrices}
In most applications, the precision matrices $\bOmega_1$ and $\bOmega
_2$ are unknown and need to 
be estimated. There is a large body of literature on estimating
precision matrices. See, for example, \cite
{bickel2008regularized,cai2011constrained,friedman2008sparse,rothman2008sparse,yuan2010high,yuan2007model,zhang2014sparse},
among others. These
methods share a common assumption that the underlying true precision
matrix is sparse. In this paper, we focus on the family of
$K_p$-sparse precision matrices as introduced in Condition \ref{conv}.
For the estimation, we use the following class of estimators.

\begin{definition}[(Acceptable estimator)]
A $p\times p$ symmetric matrix $\widehat{\bOmega}$ is an acceptable
estimator of the $K_p$-sparse population precision matrix $\bOmega$ if
it satisfies the following two conditions: (1) it is independent of the
test data and is $K_p'$-sparse with $K_p'$ a sequence of positive
integers depending only on $p$, and
(2) it satisfies the entry-wise estimation error bound $ \|\widehat
{\bOmega} - \bOmega\|_{\max} \leq C_1 K_p^2 \sqrt{(\log p)/n}$ with
some positive constant $C_1$.
\end{definition}

The same class of estimators has been introduced in and used in \cite
{fan2013optimal}. As discussed in \cite{fan2013optimal}, many existing
precision matrix estimators such as CLIME \cite{cai2011constrained} and
Glasso \cite{friedman2008sparse} are acceptable under some regularity
conditions. Other methods for estimating precision matrices can also
yield acceptable estimators under certain conditions; see \cite
{fan2013optimal} for more discussions on acceptable estimators.

For each $k=1,2$, given an acceptable estimator $\widehat{\bOmega}_k$
of $\bOmega_k$, our IIS approach transforms the data matrix as $\bZ
\widehat{\bOmega}_k$.
Similar to the last subsection, we only discuss in detail IIS based
on the transformation $\bZ\widehat{\bOmega}_1$. Then the corresponding
test statistic $\widehat{D}_j$ is
%
\begin{equation}
\label{eq: Dj-hat} \widehat{D}_j = \log\hat{\sigma}^2_j
- \sum_{k = 1}^2 (n_{k}/n) \log
\bigl[\bigl(\hat {\sigma}_j^{(k)}\bigr)^2 \bigr],
\end{equation}
where $\hat{\sigma}_j^2$ is the pooled sample variance estimate for the
$j$th\vspace*{-2pt} feature after the transformation $\bZ\widehat{\bOmega}_1$, and
$(\hat{\sigma}_j^{(k)})^2$ is the class $k$ sample variance estimate
for the $j$th feature after the transformation for $k=1,2$.

With an acceptable estimate $\widehat{\bOmega}_1$ of $\bOmega_1$, the
transformed data matrix $\bZ\widehat{\bOmega}_1$ is expected to be
close to the data matrix $\bZ\bOmega_1$. Correspondingly, the test
statistics $\widehat{D}_j$ are expected to be close to the test
statistics $\widetilde{D}_j$ defined in \eqref{eq: Dj}, which ensures
that the same selection consistency property discussed in Proposition~\ref{Prop1}
is inherited by using test statistics $\widehat{D}_j$. This
result is formally summarized below in Theorem~\ref{Thm2}.

%


Define $\widehat{\mathcal{A}}_1 = \{1\leq j \leq p\dvtx \widehat{D}_j >
\omega_n\}$ with $\omega_n>0$ the threshold level depending only on
$n$. Let
\[
T_{n,p} = \tilde{C}_1 \tilde{\tau}_{1,p}^{-1}
\tau_{2,p} \bigl(K_p + K_p'\bigr)
K_p^3 \sqrt{(\log p)/n} \max\bigl\{\bigl(K_p
+ K_p'\bigr) K_p \sqrt{(\log p)/n}, 1\bigr
\},
\]
where $\tilde{C}_1$ is some positive constant, and $\tilde\tau_{1,p}$
and $\tau_{2,p}$ are the same as in Proposition~\ref{Prop1}.



\begin{theorem}\label{Thm2} Assume that the conditions in Proposition~\ref{Prop1} are satisfied and that for each $k = 1,2$, $\widehat
{\bOmega
}_k$ is an acceptable estimator of the true precision matrix~$\bOmega
_k$. In addition, assume that Condition \ref{conv} is satisfied and
$T_{n,p} = o(n^{-\kappa})$. Then with probability at least $1 - \exp\{-
C n^{1 - 2\kappa}/[\tilde{\tau}_{1,p}^{-2} + \log^2(\tilde{\tau
}_{2,p})]\}$ for some positive constant $C$, it holds that
\[
\widehat{\mathcal{A}}_1 = \mathcal{A}_1 \qquad\mbox{with }
\omega_n \in (\alpha_n -\beta_n, \alpha_n)
\]
for large enough $n$,
where $\alpha_n = 2c n^{-\kappa} - T_{n,p}$ and $\beta_n = c
n^{-\kappa
} - 2T_{n,p}$ with $c$ defined in Condition \ref{condistin}. The same
result holds for sets $\mathcal{A}_2$ and $\widehat{\mathcal{A}}_2$
with $\widehat{\mathcal{A}}_2$ defined analogously to $\widehat
{\mathcal
{A}}_1$ using the test statistics calculated with data transformed by
$\widehat{\bOmega}_2$.
\end{theorem}

As shown in the proof of Theorem~\ref{Thm2} in 
Section~\ref{sec: SecA}, it holds that
\[
\min_{j \in\mathcal{A}_1} \widehat{D}_j \geq\alpha_n
\quad\mbox{and} \quad\max_{j \in\mathcal{A}_1^c} \widehat{D}_j \leq
\alpha_n - \beta_n,
\]
with asymptotic probability one.
The term $\beta_n$ measures how different the test statistics are in
and outside of set $\mathcal{A}_1\cup\mathcal{A}_2$. Thus, by
thresholding the test statistics $\widehat D_j$ with appropriately
selected threshold level, the index set $\mathcal{A}_1\cup\mathcal
{A}_2$ can be identified with asymptotic probability one, and
consequently, our IIS method enjoys the variable selection consistency
as described in Theorem~\ref{Thm2}. We will discuss the implementation
of IIS with test statistics \eqref{eq: Dj-hat} in detail in
Section~\ref{sec: Numerical}.

Compared to Proposition~\ref{Prop1}, the lower bound of the test
statistics over $\mathcal{A}_1$, which is given by $\alpha_n$, is
smaller than the one in Proposition~\ref{Prop1}, reflecting the
sacrifice caused by estimating precision matrices. 
The additional assumption on $T_{n,p}$ is related to the sparsity level
and estimation errors of precision matrices. Under these two
assumptions, $\alpha_n$ and $\beta_n$ are close to $2c n^{-\kappa}$ and
$c n^{-\kappa}$, the bounds given in Proposition~\ref{Prop1},
respectively, implying a relatively small price paid in estimating
precision matrices.

\section{Post-screening variable selection} \label{sec: Selection}

Denote by $\widehat{\mathcal I} = \widehat{\mathcal A}_1 \cup
\widehat
{\mathcal A}_2$ the index set identified by the IIS approach. Let $d =
|\widehat{\mathcal I}|$ be its cardinality. Then the variable selection
consistency of IIS guarantees that $\widehat{\mathcal I}$ is the true
set of interaction variables $\mathcal I$ with asymptotic probability
one. By the sparsity of $\bOmega$ assumed in Section~\ref{sec: Model},
the cardinality $d$ is equal to $q=o(\min\{n,p\})$ with overwhelming
probability. With selected variables in $\widehat{\mathcal I}$,
interactions can be reconstructed as $\mathcal B = \{Z_jZ_\ell, \mbox{
for all } j,\ell\in\widehat{\mathcal I} \}$, which indicates that IIS
reduces the dimensionality of interactions from $O(p^2)$ to less than
$o(\min\{n^2,p^2\})$ with overwhelming probability. Important questions
are how to further select active interactions and how to conduct
classification using these selected interactions.

In the classification literature, variable selection techniques have
been frequently used to construct high-dimensional classifiers, for
example, the penalized logistic regression \cite
{friedman2010regularization,zhu2004classification},
the LPD rule \cite{cai2011direct}, and the DSDA approach \cite
{mai2012direct}, among many others. In this paper, we use the idea of
penalized logistic regression to further select important main effects
and interactions. Before going into details, we first introduce some
nation. For a feature vector ${\mathbf z}=(Z_1, \ldots, Z_p)^T$, 
let ${\mathbf x}=(1, Z_1, \ldots, Z_p, Z_{1}^2, Z_{1}Z_{2}, \ldots,
Z_{p-1}Z_{p}, Z_{p}^2)^T$ be the $\tilde{p}$-dimensional full augmented
feature vector with $\tilde{p}=(p+1)(p+2)/2$.
Assume that the conditional probability of success $\pi({\mathbf
x})=P(\Delta
=1|{\mathbf x})=P(\Delta=1|{\mathbf z})$ is linked to the feature
vector ${\mathbf x}$ by the
following logistic regression model:
%
\begin{equation}
\label{def:logistic} \operatorname{logit}\bigl(\pi({\mathbf x})\bigr) = \log\frac{\pi({\mathbf x})}{1-\pi
({\mathbf x})} =
{\mathbf x} ^T\btheta,
\end{equation}
where $\btheta$ is the regression coefficient vector. Based on \eqref
{def:logistic}, a new observation ${\mathbf z}$ is classified into
class 1 if
and only if 
${\mathbf x}^T\btheta>0$.
We remark that if both ${\mathbf z}^{(1)}$ and ${\mathbf z}^{(2)}$ are Gaussian
distributed, the decision rule derived from the logistic regression
model \eqref{def:logistic} is identical to 
the Bayes rule \eqref{eq:Qfun-Bayes}, which is our main reason of using
penalized logistic regression for selecting important main effects and
interactions.

Write $\mathscr{T}\subset\{1, \ldots, \tilde{p}\}$, the set of indices
formed by the intercept, all main effects $Z_1, \ldots, Z_p$ and
interactions $Z_{k}Z_{\ell}$ with $k, \ell\in\widehat{\mathcal{I}}$.
If there is no interaction screening and $\widehat{\mathcal{I}}=\{1,
\ldots, p\}$, then $\mathscr{T}=\{1, \ldots, \tilde{p}\}$, meaning that
all pairwise interactions are used in post-screening variable selection step.

Denote by $\bX=({\mathbf x}_1, \ldots, {\mathbf x}_n)^T=(\tilde
{{\mathbf x}}_1, \ldots, \tilde
{{\mathbf x}}_{\tilde{p}})$ the full augmented design matrix with
${\mathbf x}_i$ the
full augmented feature vector for the $i$th observation ${\mathbf
z}_i$. In
order to estimate
the regression coefficient vector $\btheta$,
we consider the reduced feature space spanned by the $1+p + d(d+1)/2$
columns of $\bX$ with indices in $\mathscr{T}$ and estimate $\btheta$
by solving the following regularization problem:
%
\begin{equation}
\label{eq: elastic-obj} \widehat{\btheta} = \argmin_{\btheta\in\mathbb{R}^{\tilde{p}},
\btheta
_{\mathscr{T}^c}=\bzero} \Biggl
\{n^{-1}\sum_{i=1}^n\ell\bigl({
\mathbf x}_i^T\btheta, \Delta_i\bigr)+
\operatorname{pen}(\btheta) \Biggr\},
\end{equation}
where $\mathscr{T}^c$ is the complement of $\mathscr{T}$, $\ell
({\mathbf x}^T
\btheta, \Delta)=-\Delta({\mathbf x}^T\btheta) + \log[1+\exp
({\mathbf x}^T\btheta)]$ is
the logistic loss function and $\operatorname{pen}(\btheta)$ is some penalty
function on the parameter vector $\btheta$.
Various penalty functions have been proposed in the literature for
high-dimensional variable selection; see, for example, Lasso \cite
{tibshirani1996regression}, SCAD \cite{fan2001variable}, SICA \cite
{lv2009unified} and MCP \cite{zhang2010nearly}, among many others. See
also \cite{fan2013asymptotic} for the asymptotic equivalence of various
regularization methods.
Due to the existence of interactions, the design matrix $\bX$ can have
highly correlated columns. To overcome the difficulty caused by
potential high collinearity, in our application we propose to use the
elastic net penalty \cite{zou2005regularization}, which takes the form
$\operatorname{pen}(\btheta)=\lambda_1\|\btheta\|_1+\lambda_2\|\btheta\|_2^2$
with $\lambda_1$ and $\lambda_2$, two nonnegative regularization
parameters. Similar types of penalty functions have also been used and
studied in \cite{bunea2008honest} and \cite{hebiri2011smooth}. Note
that solving the regularization problem \eqref{eq: elastic-obj} in the
reduced parameter space $\mathscr{T}$ is computationally more efficient
than solving it in the original $\tilde p$-dimensional parameter space.

Generally speaking, the post-screening variable selection is able to
reduce the number of false positive interactions. Thus, only when there
are interactions surviving both the screening step and variable
selection step, sparse QDA will be used for classification; otherwise,
sparse LDA will be used for classification. In this sense, our approach
is adaptive and automatically chooses between sparse LDA and sparse QDA.

\subsection{Oracle inequalities}


Denote by $S = \supp(\btheta_0)$ the support of the true regression
coefficient vector $\btheta_0$ and $S^c$ its complement. Let $s=|S|$ be
the cardinality of the set $S$. For any $\bdelta= (\delta_1,\ldots,
\delta_{\tilde p})^T\in\mathbb{R}^{\tilde{p}}$, we use $\bdelta
_S$ to
denote the subvector formed by the components $\delta_j$ with $j\in S$.
The following conditions are needed for establishing the oracle
inequalities for $\widehat{\btheta}$ defined in \eqref{eq: elastic-obj}:


\begin{condition}\label{con: mean}
There exists some positive constant $0<\pi_{\min}<1/2$ such that $\pi
_{\min}<P(\Delta=1|{\mathbf z})<1-\pi_{\min}$ for all ${\mathbf z}$.
\end{condition}

\begin{condition}\label{con: RE-population}
There exists some constant $\phi>0$ such that
%
\begin{equation}
\label{eq: RE} 
\bdelta^T\tilde{\bSig}\bdelta\geq
\phi^2\bdelta_S^T\bdelta_S
\end{equation}
for any $\bdelta\in\mathbb{R}^{\tilde{p}}$ satisfying $\|\bdelta
_{S^c}\|
_1\leq4(s^{1/2}+\lambda_1^{-1}\lambda_2\|\btheta_0\|_2)\|\bdelta
_{S}\|_2$,
where $\tilde{\bSig}=E({\mathbf x}^T{\mathbf x})$.
\end{condition}


Condition \ref{con: mean} is a mild condition which is commonly imposed
in logistic regression and ensures that the conditional variance of the
response variable is uniformly bounded away from zero. Condition \ref
{con: RE-population} is inspired by the restricted eigenvalue (RE)
assumptions in \cite{bickel2009simultaneous}, where it was introduced
for establishing the oracle inequalities for the lasso estimator \cite
{tibshirani1996regression} and the Dantzig selector \cite
{candes2007dantzig}. 
The set on which \eqref{eq: RE} holds in Condition \ref{con:
RE-population} also involves $\lambda_1^{-1}\lambda_2\|\btheta_0\|_2$,
which is needed to deal with the $L_2$ term in the elastic net penalty
\cite{zou2005regularization}.
A similar condition has been used in \cite{hebiri2011smooth} for
studying the oracle inequalities for the smooth-Lasso and other $\ell
_1+\ell_2$ methods in ultrahigh-dimensional linear regression models
with deterministic design and no interactions. In our setting, the
logistic regression model with random design and the existence of
interactions add extra technical difficulties in establishing the
oracle inequalities. 

\begin{theorem}\label{thmm: oracle}
Assume that all conditions in Theorem~\ref{Thm2} and Conditions
\ref
{con: mean}--\ref{con: RE-population} are satisfied. 
Moreover, assume that $\lambda_1\geq c_0\sqrt{\log(p)/n}$ with some
positive constant $c_0$, $5s^{1/2}+4\lambda_1^{-1}\lambda_2\|\btheta
_0\|
_2=O(n^{\xi/2})$, and $\log(p)=o(n^{1/2-2\xi})$ with some constant
$0\leq\xi<1/4$. Then with
probability at least $1-\break \exp\{- C n^{1 - 2\kappa}/[\tilde{\tau
}_{1,p}^{-2} + \log^2(\tilde{\tau}_{2,p})]\}-O(p^{-c_1})$, it holds
simultaneously that
%
\begin{eqnarray*}
\|\widehat{\btheta}-\btheta_0\|_1& \leq& 32
\widetilde{C}^{-1}\phi^{-2}\bigl(\lambda_1s^{1/2}+
\lambda_2\| \btheta_0\|_2
\bigr)^2/\lambda_1,
\\
n^{-1/2}\bigl\|\bX(\widehat{\btheta}-\btheta_0)
\bigr\|_2 &\leq& 4\widetilde{C}^{-1}\phi^{-1}\bigl(
\lambda_1s^{1/2}+\lambda_2\|
\btheta_0\|_2\bigr), 
\end{eqnarray*}
where $\widetilde{C}$ is some positive constant. Moreover, the same
results hold with probability at least $1-O(p^{-c_1})$ for the
regularized estimator $\widehat{\btheta}$ 
without the interaction screening step, that is, without the constraint
$\btheta_{\mathscr{T}^c}=\bzero$ in \eqref{eq: elastic-obj}.
\end{theorem}

Theorem~\ref{thmm: oracle} presents the oracle inequalities for the
regularized estimator $\widehat{\btheta}$ defined in \eqref{eq:
elastic-obj}. It extends the oracle inequalities in Theorem~1 of \cite
{hebiri2011smooth} from the linear model with deterministic design and
no interactions to the logistic regression model with random design and
interactions. Dealing with interactions and large random design matrix
needs more delicate analysis. It is worth pointing out that the results
in Theorem~\ref{thmm: oracle} also apply to the regularized estimator
with $d=p$, that is, the case without interaction screening.

\subsection{Oracle inequality for misclassification rate}

Recall that based on the logistic regression model \eqref
{def:logistic}, the oracle classifier classifies a new observation
${\mathbf z}
$ to class $1$ if and only if
${\mathbf x}^T\btheta_0>0$,
where ${\mathbf x}$ is the $\tilde{p}$-dimensional augmented feature vector
corresponding to ${\mathbf z}$. Thus the oracle misclassification rate is
\[
R=\pi R(2|1) + (1-\pi) R(1|2),
\]
where $R(i|j)$ is the probability that a new observation from class $j$
is misclassified to class $i$ based on the oracle classifier. As
discussed in the last subsection, the oracle classifier 
${\mathbf x}^T\btheta_0$ is the Bayes rule if the feature vectors
${\mathbf z}^{(1)}$
and ${\mathbf z}^{(2)}$ from classes $1$ and $2$ are both Gaussian.

Correspondingly, given the sample $\{({\mathbf z}_i^T, \Delta_i)\}_{i=1}^n$,
the misclassification rate of the plug-in classifier 
${\mathbf x}^T\widehat{\btheta}$ with $\widehat{\btheta}$ defined
in \eqref{eq:
elastic-obj} takes the following form:
\[
R_n=\pi R_n(2|1) + (1-\pi) R_n(1|2),
\]
where $R_n(i|j)$ is the probability that a new observation from class
$j$ is misclassified to class $i$ by the plug-in classifier.

We introduce some notation before stating our theoretical result on
misclassification rate. Denote by $F_1(x)$ and $F_2(x)$ the cumulative
distribution functions of the oracle classifier 
${\mathbf x}^T\btheta_0$ under classes $1$ and $2$, respectively. Let
\[
r_n=\max\Bigl\{\sup_{x\in[-\epsilon_0, \epsilon_0]}\bigl|F_1'(x)\bigr|,
\sup_{x\in
[-\epsilon_0, \epsilon_0]}\bigl|F_2'(x)\bigr|\Bigr\},
\]
where $\epsilon_0$ is a small positive constant, and $F_1'(x)$ and
$F_2'(x)$ are the first-order derivatives of $F_1(x)$ and $F_2(x)$,
respectively.

\begin{condition}\label{con: CDF} Let 
$\Lambda_n=\log(p) (\lambda_1s^{1/2}+\lambda_2\|\btheta_0\|
_2)^2/\lambda
_1$. It holds that
$\Lambda_n=o(1)$ and $r_n\Lambda_n=o(1)$.
\end{condition}

\begin{theorem}\label{thmm: MR}
Assume that all conditions in Theorem~\ref{thmm: oracle} and Condition
\ref{con: CDF} are satisfied. 
Then with probability at least $1-\exp\{- C n^{1 - 2\kappa}/[\tilde
{\tau
}_{1,p}^{-2} + \break \log^2(\tilde{\tau}_{2,p})]\}-O(p^{-c_1})$, we have
%
\begin{equation}
\label{eq:MR} 0\leq R_n\leq R+O\bigl(p^{-c_1}\bigr) +
O(r_n\Lambda_n)
\end{equation}
for all sufficiently large $n$, where $c_1$ is some positive constant.
Moreover, the same inequality holds with probability at least
$1-O(p^{-c_1})$ for the plug-in classifier based on the regularization
estimator $\widehat{\btheta}$ without interaction screening.
\end{theorem}

Theorem~\ref{thmm: MR} ensures that with overwhelming probability, the
misclassification rate of the plug-in classifier is at most
$O(p^{-c_1}) + O(r_n\Lambda_n)$ worse than that of the oracle
classifier. If $r_n$ is upper-bounded by some constant, $\lambda_1 =
O(\sqrt{(\log p)/n})$, and 
$\lambda_2\|\btheta_0\|_2 = O(s^{1/2}\lambda_1)$, then \eqref{eq:MR}
becomes $0\leq R_n \leq R+O(p^{-c_1}) + O(s(\log p)^{3/2}n^{-1/2})$. In
the setting of two-class Gaussian classification, the misclassification
rate $R_n$ can also be lower bounded by $R$, by noting that the oracle
classifier 
${\mathbf x}^T\btheta_0$ is the Bayes rule. Thus the plug-in
classifier is
consistent. This result is formally summarized in the following corollary.

\begin{corollary}
Assume that both ${\mathbf z}^{(1)}$ and ${\mathbf z}^{(2)}$ are
Gaussian distributed.
Then under the same conditions as in Theorem~\ref{thmm: MR}, with
probability at least $1-\exp\{- C n^{1 - 2\kappa}/[\tilde{\tau
}_{1,p}^{-2} + \log^2(\tilde{\tau}_{2,p})]\}-O(p^{-c_1})$, it holds that
\[
R \leq R_n \leq R + O\bigl(p^{-c_1}\bigr) +
O(r_n\Lambda_n).\vadjust{\goodbreak}
\]
\end{corollary}

\section{Numerical studies}\label{sec: Numerical}

\subsection{Implementation} \label{sec: Implement}

We apply the SIRI method in \cite{jiang2014sliced} to implement IIS in
our proposal. See Section~5.2 in \cite{jiang2014sliced} for more
details on how to choose thresholds in SIRI. The R code for SIRI is
available at \url{http://www.people.fas.harvard.edu/\textasciitilde junliu/SIRI/}.

It is worth mentioning that as recommended in \cite{jiang2014sliced},
SIRI is implemented as an iterative stepwise procedure. That is, the
next active interaction variable is chosen based on the current set of
interaction variables rather than using a one-time hard-thresholding to
select all interaction variables. The iterative stepwise procedure is
more stable in practice. Jiang and Liu \cite{jiang2014sliced} proved
the nice property of SIRI method in selecting interaction variables in
the sliced inverse index model setting. We remark that the same
theoretical results hold under our model setting as long as an extra
condition similar to the stepwise detectable condition in \cite
{jiang2014sliced} is imposed on the population variances.
Since the proofs are very similar to the ones in \cite{jiang2014sliced},
to save space, we do not formally state the results here. Instead, we
refer the readers to \cite{jiang2014sliced} for more details.

In the second stage of our proposal, we employ the R package glmnet for
variable selection. An refitting step after selection is added when
calculating classification error. For the ease of presentation, our
two-stage procedure is referred to as IIS-SQDA.

For comparisons, we also include LDA, QDA, penalized logistic
regression (PLR), DSDA and the oracle procedure (Oracle). The LDA and
QDA methods are implemented by directly plugging in the sample
estimates of the unknown parameters.
The oracle procedure uses the information of the true underlying
sparse model and is thus a low-dimensional QDA. For PLR, we consider
two different versions, PLR and PLR2, where the former uses main
effects only and the latter includes additionally all possible pairwise
interactions. For fair comparison, an refitting step is also conducted
for PLR and PLR2, as we do for IIS-SQDA.

\subsection{Simulation studies} \label{sec: Simulation}

We conducted two simulation studies to evalaute the performace of
IIS-SQDA. The class $1$ distribution is chosen to be $N(\bmu_1,\break  \bSig
_1)$ with $\bmu_1=\bSig_1\bdelta$ and $\bSig_1=\bOmega_1^{-1}$,
and the
class 2 distribution is chosen to be $N(\bzero, \bSig_2)$ with $\bSig
_2=\bOmega_2^{-1}$, where $\bOmega_1$, $\bOmega_2$ and $\bdelta$ will
be specified later.

\subsubsection{Study 1} \label{sec: Study-true-Omega}

We demonstrate the performance of the oracle-assisted IIS approach and
examine the resulting classification and variable selection
performance. The results presented here can be used as a benchmark for
evaluating the performance of IIS with unknown precision matrices. We
consider the following setting for $\bdelta$ and precision matrices
$\bOmega_1$ and $\bOmega_2$:
\begin{longlist}
\item[Model 1:] $(\bOmega_1)_{ij}=0.5^{|i-j|}$, $\bOmega
_2=\bOmega_1+\bOmega$ where $\bOmega$ is a symmetric and sparse matrix
with $\bOmega_{5, 5}=\bOmega_{25, 25}=\bOmega_{45, 45}=-0.29$ and
$\bOmega_{5, 25}=\bOmega_{5, 45}=\break \bOmega_{25, 45}=-0.15$. The other 3
nonzero entries in the lower triangle of $\bOmega$ are determined by
symmetry. $\bdelta=(0.6, 0.8, 0, \ldots, 0)^T$.
The dimension $p$ is $2000$.
\end{longlist}
Thus there are two main effects and six interaction terms under our
broad definition of interaction in the Bayes rule \eqref{eq:Qfun-Bayes}.

We use two performance measures, false positive (FP), and false
negative (FN), to evaluate the screening performance of IIS. FP is
defined as the number of irrelevant interaction variables falsely kept
while FN is defined as the number of true interaction variables falsely
excluded by IIS. An effective variable screening procedure is expected
to have the value of FP reasonably small and the value of FN close to
zero. The former implies that the variable screening procedure can
effectively reduce the dimensionality whereas the latter implies that
the sure screening property holds. The means and standard errors (in
parentheses) of FP and FN for interaction variables based on 100
replications are $0.63\ (0.08)$ and $0.14\ (0.03)$, respectively, in
the screening step. This demonstrates the fine performance of our IIS
approach in selecting interaction variables.

We further investigate the classification and variable selection
performance of our proposal. Five performance measures are employed to
summarize the results. The first measure is the misclassification rate
(MR), which is calculated as the proportion of observations in an
independently simulated test set of size 10,000 being allocated to the
incorrect class. The second and third are FP.main and FP.inter, which
represent the numbers of irrelevant main effects and irrelevant
interaction effects falsely included in the classification rule,
respectively. The fourth and fifth are FN.main and FN.inter, which
represent the numbers of relevant main effects and relevant interaction
effects falsely excluded in the classification rule, respectively. Note
that the definitions of FP.inter and FN.inter here are different from
those screening performance measures FP and FN, which are defined
earlier. In fact, FP.inter and FN.inter are defined with respect to the
number of interaction effects whereas screening performance measures FP
and FN are defined with respect to the number of interaction variables.

The variable selection and classification results for different methods
are reported in Table~\ref{table: Study-true-Omega}.
PLR2 is not computationally efficient in this case due to the huge
number of two-way interactions. The conventional LDA and QDA are not
applicable as $n_1=n_2=100<p$. So we only compare the variable
selection and classification performance of our proposal, IIS-SQDA,
with DSDA, PLR and the Oracle.
It is made clear that IIS-SQDA has better classification performance
than PLR and DSDA.

\begin{table}
\caption{The means and standard errors (in parentheses) of various
performance measures by different classification methods for study 1
based on 100 replications}
\label{table: Study-true-Omega}
\begin{tabular*}{\textwidth}{@{\extracolsep{\fill}}lcccc@{}}
\hline
\textbf{Measure} & \textbf{PLR} & \textbf{DSDA} & \textbf{IIS-SQDA} & \textbf{Oracle} \\
\hline
MR (\%) & 40.59 (0.40) & 38.04 (0.35) & 15.09 (0.40) & 12.07 (0.06) \\
FP.main & 25.69 (5.45) & 22.80 (5.10) & 2.63 (0.66) & 0 (0) \\
FP.inter & -- & -- & 0.62 (0.12) & 0 (0) \\
FN.main & 1.80 (0.04) & 0.82 (0.05) & 1.22 (0.05) & 0 (0) \\
FN.inter & -- & -- & 0.47 (0.10) & 0 (0) \\
\hline
\end{tabular*}
\end{table}

\subsubsection{Study 2} \label{sec: Study-estimated-Omega}
In this study, we evaluate the performance of the IIS approach with the
estimated precision matrices and examine the resulting classification
and variable selection performance. We consider the following four
different model settings for precision matrices:
\begin{longlist}[Model 5:]
\item[Model 2:] $\bOmega_1=I_p$, $\bOmega_2=\bOmega_1+\bOmega$,
where $\bOmega$ is a symmetric and sparse matrix with $\bOmega_{10,
10}=\bOmega_{30, 30}=\bOmega_{50, 50}=-0.6$ and $\bOmega_{10,
30}=\bOmega_{10, 50}=\bOmega_{30, 50}=-0.15$. The other 3 nonzero
entries in the lower triangle of $\bOmega$ are determined by symmetry.
$\bdelta=(0.6, 0.8, 0, \ldots, 0)^T$.
\item[Model 3:] $\bOmega_1$ is a band matrix with $(\bOmega
_1)_{ii}=1$ for $i=1,\ldots, p$ and $(\bOmega_1)_{ij}=0.3$ for $|i-j|=1$.
$\bOmega_2=\bOmega_1+\bOmega$ where $\bOmega$ is a symmetric and
sparse matrix with $\bOmega_{10, 10}=-0.3785$, $\bOmega_{10,
30}=0.0616$, $\bOmega_{10, 50}=0.2037$, $\bOmega_{30, 30}=-0.5482$,
$\bOmega_{30, 50}=0.0286$ and $\bOmega_{50, 50}=-0.4614$. The other 3
nonzero entries in the lower triangle of $\bOmega$ are determined by
symmetry. $\bdelta=(0.6, 0.8, 0, \ldots, 0)^T$.
\item[Model 4:] Similar to model 1 in the last subsection,
except for the dimension $p$.
\item[Model 5:] $\bOmega_1$ is a block diagonal matrix comprised
of equal blocks $\bA$, where $\bA$ is a $2$-by-$2$ matrix with diagonal
elements equal to $1$ and off-diagonal elements equal to $0.4$.
$\bOmega
_2=\bOmega_1+\bOmega$ where $\bOmega$ is a symmetric and sparse matrix
with $\bOmega_{3, 3}=\bOmega_{6, 6}=\bOmega_{9, 9}=\bOmega_{12,
12}=-0.2$, $\bOmega_{3, 6}=\bOmega_{9, 12}=0.4$ and $\bOmega_{3,
9}=\bOmega_{3, 12}=\bOmega_{6, 9}=\bOmega_{6, 12}=-0.4$. The other $6$
nonzero entries in the lower triangle of $\bOmega$ are determined by
symmetry. The nonzero elements of $\bdelta$ are located at coordinates
$3$, $6$, $9$ and $12$. The corresponding values for these nonzero
elements are simulated from a uniform distribution over $[0.3,0.7]$ and
remain unchanged during simulations.
\end{longlist}
For each model, we consider three different dimentionalities, $p=50$,
$p=200$ and $p=500$.
There are two main effects and six interaction terms (including
quadratic terms) in the Bayes rules for models 2--4,
four main effects and ten interaction terms in the Bayes rules for
model 5.
In models 2--4 no interaction variables are main effect variables
whereas in model 5 all interaction variables are also main effect variables.

We use the same measures as in study 1 to examine the variable
screening performance of the IIS approach and the variable selection
and classification performance of IIS-SQDA. The means and standard
errors (in parentheses) of FP and FN for these models based on 100
replications are reported in Table~\ref{table: screen-estimated-Omega},
which shows the effectiveness of our interaction screening approach.
For comparison purposes, we also include in Table~\ref{table:
screen-estimated-Omega} the screening results by oracle-assisted IIS.
It is interesting to observe that the IIS with estimated precision
matrices gives smaller FNs than and comparable FPs to the IIS with true
precision matrices. 

\begin{table}
\caption{Interaction screening results for models 2--5. The numbers
reported are the means and standard errors (in parentheses) of FP and
FN based on 100 replications}
\label{table: screen-estimated-Omega}
\begin{tabular*}{\textwidth}{@{\extracolsep{4in minus 4in}}lccccc@{}}
\hline
& &\multicolumn{2}{c}{\textbf{IIS with true} $\bolds{\Omega_1}$ \textbf{and}
$\bolds{\Omega_2}$}
&\multicolumn{2}{c@{}}{\textbf{IIS with estimated} $\bolds{\Omega_1}$ \textbf{and} $\bolds{\Omega_2}$} \\[-6pt]
& &\multicolumn{2}{c}{\hrulefill}
&\multicolumn{2}{c@{}}{\hrulefill} \\
$\bolds{p}$ & \textbf{Model} & \textbf{FP} & \textbf{FN} & \textbf{FP} &
\multicolumn{1}{c@{}}{\textbf{FN}} \\
\hline
\phantom{0}50 & Model 2 & 0.45 (0.08) & 0.02 (0.01) & 1.57 (0.15) & 0.01 (0.01) \\
& Model 3 & 0.86 (0.09) & 0.48 (0.06) & 1.93 (0.15) & 0.15 (0.04) \\
& Model 4 & 1.68 (0.13) & 0.09 (0.03) & 1.04 (0.11) & 0.01 (0.01) \\
& Model 5 & 1.79 (0.16) & 0.02 (0.02) & 1.54 (0.13) & 0.01 (0.01) \\[3pt]
200 & Model 2 & 0.43 (0.08) & 0.04 (0.02) & 1.16 (0.13) & 0.02 (0.01) \\
& Model 3 & 0.74 (0.09) & 0.48 (0.05) & 1.03 (0.14) & 0.15 (0.04) \\
& Model 4 & 1.52 (0.12) & 0.08 (0.03) & 0.44 (0.07) & 0.03 (0.02) \\
& Model 5 & 1.10 (0.12) & 0.36 (0.08) & 0.90 (0.10) & 0.04 (0.02) \\[3pt]
500 & Model 2 & 0.42 (0.07) & 0.11 (0.03) & 0.68 (0.09) & 0.01 (0.01) \\
& Model 3 & 0.53 (0.06) & 0.73 (0.07) & 0.65 (0.09) & 0.21 (0.04) \\
& Model 4 & 1.25 (0.12) & 0.09 (0.03) & 0.43 (0.07) & 0.03 (0.02) \\
& Model 5 & 0.85 (0.10) & 0.42 (0.09) & 0.59 (0.09) & 0.03 (0.02) \\
\hline
\end{tabular*}
\end{table}

Tables~\ref{table: select-M2}--\ref{table: select-M5} summarize the
variable selection and classification results based on 100
replications. 
We observe the following:
\begin{longlist}[(1)]
\item[(1)] IIS-SQDA exhibits the best performance in terms of MR and
interaction selection across all settings.

\item[(2)] PLR2 also has good classification accuracy in low-dimensional
situations ($p=50$), but it has inferior interaction selection results
than IIS-SQDA in all settings.

\item[(3)] All linear classifiers have poor performance when the true
classification boundary is nonlinear.

\item[(4)] Comparing QDA with LDA shows that including all possible
interactions may not necessarily improve the classification
performance. This is not surprising because QDA has many more
parameters to estimate than LDA, while the sample size is very limited.
Thus interaction selection is very important, even with moderate dimensionality.

\item[(5)] Comparing the results of QDA with those of PLR2 or IIS-SQDA, we
observe that the classification performance can be improved
substantially by using interaction screening and selection.
Particularly, in most cases, the improvement becomes more significant
as the dimensionality increases.
\end{longlist}

Another phenomenon we observed in simulation is that when 
the number of predictors $p$ is as high as $p=500$, PLR2 requires a
huge memory space that it easily causes memory outflow in a regular
office PC with 8~GB memory. 

In addition, note that the misclassification rates of all methods in
model 5 are significantly higher than that of the Oracle classifier. We
emphasize that it is due to the small true coefficients in the Bayes
rule and the relatively complex true model. In fact, the setting of
model 5 is so challenging that all other methods have close to or over
40\% MR when $p=200$ or $500$.

\begin{table}
\caption{The means and standard errors (in parentheses) of various
performance measures by different classification methods for model 2
based on 100 replications}\label{table: select-M2}
\begin{tabular*}{\textwidth}{@{\extracolsep{\fill}}lcccccc@{}}
\hline
$\bolds{p}$ & \textbf{Method} & \textbf{MR (\%)} & \textbf{FP.main} & \textbf{FP.inter} & \textbf{FN.main} & \multicolumn{1}{c@{}}{\textbf{FN.inter}} \\
\hline
\phantom{0}50 & LDA & 37.91 (0.13) & 48 (0) & -- & 0 (0) & --\\
& QDA & 39.89 (0.11) & 48 (0) & 1269 (0) & 0 (0) & 0 (0) \\
& PLR & 32.83 (0.23) & 2.37 (0.49) &-- &1.09 (0.03) &-- \\
& DSDA & 32.70 (0.18) & 4.61 (0.74) & -- & 0.10 (0.03) & --\\
& PLR2 & 22.56 (0.33) & 0.13 (0.05) &3.17 (0.70) & 0.35 (0.05) & 0.75
(0.09)\\
& IIS-SQDA & 21.78 (0.22) & 3.67 (0.67) &1.32 (0.23) & 0.08 (0.03) &
0.09 (0.04)\\
& Oracle & 19.86 (0.08) & 0 (0) & 0 (0) & 0 (0) & 0 (0) \\[3pt]
200 
& PLR & 33.64 (0.31) & 4.29 (1.34) &-- &1.09 (0.03) &-- \\
& DSDA & 33.33 (0.26) & 10.83 (2.25) & -- & 0.18 (0.04) & --\\
& PLR2 & 24.65 (0.51) & 0.11 (0.05) &7.71 (2.27) & 0.42 (0.06) & 0.93
(0.09)\\
& IIS-SQDA & 22.14 (0.30) & 4.48 (0.91) &0.54 (0.11) & 0.09 (0.03) &
0.15 (0.05)\\
& Oracle & 19.66 (0.06) & 0 (0) & 0 (0) & 0 (0) & 0 (0) \\[3pt]
500 
& PLR & 34.59 (0.39) & 6.00 (1.46) &-- &1.12 (0.03) &-- \\
& DSDA & 33.87 (0.28) & 14.76 (3.10) & -- & 0.17 (0.04) & --\\
& PLR2 & 26.83 (0.58) & 0.07 (0.04) &8.95 (2.02) & 0.56 (0.06) & 1.53
(0.11)\\
& IIS-SQDA & 22.09 (0.30) & 3.25 (1.02) &0.25 (0.08) & 0.25 (0.05) &
0.69 (0.09)\\
& Oracle & 19.65 (0.06) & 0 (0) & 0 (0) & 0 (0) & 0 (0) \\
\hline
\end{tabular*}
\end{table}

\begin{table}
\caption{The means and standard errors (in parentheses) of various
performance measures by different classification methods for model 3
based on 100 replications}
\label{table: select-M3}
\begin{tabular*}{\textwidth}{@{\extracolsep{\fill}}lcccccc@{}}
\hline
$\bolds{p}$ & \textbf{Method} & \textbf{MR (\%)} & \textbf{FP.main} & \textbf{FP.inter} & \textbf{FN.main} & \multicolumn{1}{c@{}}{\textbf{FN.inter}} \\
\hline
\phantom{0}50 & LDA & 39.43 (0.15) & 48 (0) & -- & 0 (0) & --\\
& QDA & 43.47 (0.10) & 48 (0) & 1269 (0) & 0 (0) & 0 (0) \\
& PLR & 36.12 (0.26) & 5.95 (0.93) &-- &1.21 (0.04) &-- \\
& DSDA & 35.05 (0.22) & 8.81 (1.06) & -- & 0.07 (0.03) & --\\
& PLR2 & 30.15 (0.44) & 0.51 (0.14) &11.26 (2.78) & 0.60 (0.05) & 2.62
(0.09)\\
& IIS-SQDA & 27.56 (0.27) & 5.60 (0.82) &2.16 (0.32) & 0.19 (0.04) &
2.05 (0.09)\\
& Oracle & 24.13 (0.07) & 0 (0) & 0 (0) & 0 (0) & 0 (0) \\[3pt]
200 
& PLR & 37.62 (0.34) & 7.82 (1.87) &-- &1.47 (0.05) &-- \\
& DSDA & 36.34 (0.30) & 15.06 (3.37) & -- & 0.36 (0.05) & --\\
& PLR2 & 32.55 (0.53) & 0.25 (0.06) &17.44 (3.63) & 0.90 (0.05) & 2.72
(0.08)\\
& IIS-SQDA & 26.94 (0.31) & 6.43 (1.24) &0.78 (0.17) & 0.42 (0.05) &
2.22 (0.08)\\
& Oracle & 22.99 (0.07) & 0 (0) & 0 (0) & 0 (0) & 0 (0) \\[3pt]
500 
& PLR & 38.82 (0.33) & 9.31 (1.99) &-- &1.58 (0.05) &-- \\
& DSDA & 37.10 (0.29) & 16.06 (3.02) & -- & 0.42 (0.05) & --\\
& PLR2 & 35.45 (0.64) & 0.34 (0.09) &55.69 (12.67) & 0.99 (0.05) &
3.05 (0.10)\\
& IIS-SQDA & 26.78 (0.31) & 3.22 (1.09) &0.23 (0.05) & 0.98 (0.02) &
2.65 (0.09)\\
& Oracle & 23.00 (0.08) & 0 (0) & 0 (0) & 0 (0) & 0 (0) \\
\hline
\end{tabular*}
%
\end{table}

\begin{table}
\caption{The means and standard errors (in parentheses) of various
performance measures by different classification methods for model 4
based on 100 replications}
\label{table: select-M4}
\begin{tabular*}{\textwidth}{@{\extracolsep{\fill}}lcccccc@{}}
\hline
$\bolds{p}$ & \textbf{Method} & \textbf{MR (\%)} & \textbf{FP.main} & \textbf{FP.inter} & \textbf{FN.main} & \multicolumn{1}{c@{}}{\textbf{FN.inter}} \\
\hline
\phantom{0}50 & LDA & 38.84 (0.16) & 48 (0) & -- & 0 (0) & --\\
& QDA & 31.10 (0.16) & 48 (0) & 1269 (0) & 0 (0) & 0 (0) \\
& PLR & 36.06 (0.24) & 5.89 (0.78) &-- &1.39 (0.05) &-- \\
& DSDA & 35.36 (0.21) & 10.41 (1.18) & -- & 0.24 (0.04) & --\\
& PLR2 & 16.55 (0.40) & 0.40 (0.08) &22.80 (1.72) & 1.08 (0.06) & 0.33
(0.06)\\
& IIS-SQDA & 15.49 (0.33) & 9.51 (1.34) &2.91 (0.38) & 0.39 (0.05) &
0.04 (0.03)\\
& Oracle & 12.14 (0.06) & 0 (0) & 0 (0) & 0 (0) & 0 (0) \\[3pt]
200 
& PLR & 38.01 (0.30) & 9.86 (2.04) &-- &1.64 (0.05) &-- \\
& DSDA & 36.39 (0.25) & 13.98 (2.18) & -- & 0.46 (0.05) & --\\
& PLR2 & 16.79 (0.48) & 0.09 (0.03) &19.99 (1.76) & 1.40 (0.05) & 0.48
(0.08)\\
& IIS-SQDA & 13.98 (0.28) & 2.30 (0.72) &0.26 (0.09) & 0.98 (0.05) &
0.10 (0.05)\\
& Oracle & 12.12 (0.07) & 0 (0) & 0 (0) & 0 (0) & 0 (0) \\[3pt]
500 
& PLR & 39.51 (0.35) & 12.98 (2.13) &-- &1.72 (0.05) &-- \\
& DSDA & 37.90 (0.29) & 24.04 (3.94) & -- & 0.53 (0.05) & --\\
& PLR2 & 16.38 (0.52) & 0.06 (0.02) &16.79 (1.36) & 1.43 (0.05) & 0.74
(0.10)\\
& IIS-SQDA & 14.10 (0.28) & 2.11 (0.57) &0.16 (0.07) & 1.07 (0.05) &
0.12 (0.06)\\
& Oracle & 12.11 (0.06) & 0 (0) & 0 (0) & 0 (0) & 0 (0) \\
\hline
\end{tabular*}
%
\end{table}

\begin{table}
\caption{The means and standard errors (in parentheses) of various
performance measures by different classification methods for model 5
based on 100 replications}
\label{table: select-M5}
\begin{tabular*}{\textwidth}{@{\extracolsep{\fill}}lcccccc@{}}
\hline
$\bolds{p}$ & \textbf{Method} & \textbf{MR (\%)} & \textbf{FP.main} & \textbf{FP.inter} & \textbf{FN.main} & \multicolumn{1}{c@{}}{\textbf{FN.inter}} \\
\hline
\phantom{0}50 & LDA & 43.18 (0.14) & 46 (0) & -- & 0 (0) & --\\
& QDA & 41.69 (0.12) & 46 (0) & 1265 (0) & 0 (0) & 0 (0) \\
& PLR & 40.16 (0.26) & 4.77 (0.73) &-- &1.93 (0.10) &-- \\
& DSDA & 38.89 (0.26) & 7.98 (1.22) & -- & 1.12 (0.10) & --\\
& PLR2 & 34.55 (0.39) & 1.06 (0.22) &19.51 (3.53) & 2.14 (0.11) & 4.16
(0.13)\\
& IIS-SQDA & 27.68 (0.23) & 7.64 (0.86) &2.11 (0.28) & 0.90 (0.09) &
2.61 (0.18)\\
& Oracle & 22.30 (0.10) & 0 (0) & 0 (0) & 0 (0) & 0 (0) \\[3pt]
200 
& PLR & 42.15 (0.32) & 18.18 (3.10) &-- &2.22 (0.12) &-- \\
& DSDA & 39.22 (0.32) & 16.23 (3.82) & -- & 1.36 (0.11) & --\\
& PLR2 & 41.50 (0.38) & 0.34 (0.08) &72.73 (10.99) & 2.66 (0.10) &
5.24 (0.14)\\
& IIS-SQDA & 30.04 (0.32) & 11.29 (1.83) &0.91 (0.18) & 1.52 (0.10) &
4.08 (0.17)\\
& Oracle & 22.24 (0.08) & 0 (0) & 0 (0) & 0 (0) & 0 (0) \\[3pt]
500 
& PLR & 43.83 (0.32) & 29.19 (4.70) &-- &2.36 (0.13) &-- \\
& DSDA & 40.03 (0.32) & 20.54 (4.30) & -- & 1.58 (0.10) & --\\
& PLR2 & 44.92 (0.32) & 0.77 (0.13) &123.39 (15.77) & 2.97 (0.09) &
7.19 (0.15)\\
& IIS-SQDA & 32.84 (0.32) & 19.59 (3.32) &0.57 (0.10) & 1.61 (0.12) &
4.61 (0.18)\\
& Oracle & 22.12 (0.07) & 0 (0) & 0 (0) & 0 (0) & 0 (0) \\
\hline
\end{tabular*}
%
\end{table}

\subsection{Real data analysis} \label{sec: Real-Data}

We apply the same classification methods as in Section~\ref{sec:
Simulation} to the breast cancer data, originally studied in \cite
{van2002gene}.
The purpose of the study is to classify female breast cancer patients
according to relapse and nonrelapse clinical outcomes using gene
expression data.
The total sample size is 78 with 44 patients in the good prognosis
group and 34 patients in the poor prognosis group.
There are some missing values with one patient in the poor prognosis
group so it was removed from study here. Thus $n_1=44$ and $n_2=33$.
Our study uses the
$p=231$ genes reported in \cite{van2002gene}.

We randomly split the 77 samples into a training
set and a test set such that the training set consists of 26 samples
from the good prognosis group and 19 samples from the poor prognosis
group. Correspondingly, the test
set has 18 samples from the good prognosis group and 14 samples from
the poor prognosis group. For each split, we applied four different
methods, PLR, PLR2, DSDA and IIS-SQDA to the training data
and then calculated the classification error using the test
data. The tuning parameters were selected using the cross-validation.
We repeated the random splitting for 100 times. The means and standard
errors of classification errors and model sizes for different
classification methods are summarized in Table~\ref{table: Veer-Data}.
The average number of genes contributing to the selected interactions
over 100 random splittings were $22.96$ and $2.86$ for PLR2 and
IIS-SQDA, respectively.
We can observe that our proposed procedure has the best classification
performance.

\begin{table}
\caption{Misclassification rate and model size on the breast cancer
data in \cite{van2002gene} over 100 random splits. Standard errors are
in the parentheses}
\label{table: Veer-Data}
\begin{tabular*}{\textwidth}{@{\extracolsep{\fill}}lcccc@{}}
\hline
 &  & \multicolumn{3}{c@{}}{\textbf{Model size}}
\\[-3pt]
 &  & \multicolumn{3}{c@{}}{\hrulefill}\\
\textbf{Method}& \multicolumn{1}{c}{\textbf{MR (\%)}}& \multicolumn{1}{c}{\textbf{Main}} & \multicolumn{1}{c}{\textbf{Interaction}} &
\multicolumn{1}{c@{}}{\textbf{All}} \\
\hline
DSDA & 23.62 (0.74) & 37.38 (1.57) & -- & 37.38 (1.57) \\
PLR & 21.72 (0.78) & 45.04 (1.35) & -- & 45.04 (1.35) \\
PLR2 & 40.47 (0.61) & 14.87 (1.81) & 19.95 (3.28) & 34.82 (4.77) \\
IIS-SQDA & 19.97 (0.77) & 47.77 (1.16) & 3.03 (0.32) & 50.80 (1.31) \\
\hline
\end{tabular*}
\end{table}

\section{Discussion} \label{sec: Discussion}

We have proposed a new two-stage procedure, IIS-SQDA, for two-class
classification with possibly unequal covariance matrices in the
high-dimensional setting. The proposed procedure first selects
interaction variables and reconstructs interactions using these
retained variables and then achieves main effects and interactions
selection through regularization. The fine performance of IIS-SQDA has
been demonstrated through theoretical study and numerical analyses.

For future study, it would be interesting to extend the proposed
procedure to multi-class classification problems.
In addition, IIS transforms the data using the CLIME estimates of the
precision matrices $\bOmega_1$ and $\bOmega_2$, which can be slow to
calculate when 
the number of predictors $p$ is very large. One possible solution is to
first reduce the dimensionality using some screening method and then
apply our IIS-SQDA for interaction screening and classification. We are
also in the process of developing a scalable version of the IIS which
significantly improves the computational efficiency.


\section{Proofs of main theorems} \label{sec: SecA}

In this section, we list main lemmas and present the proofs for main
theorems. The secondary lemmas and additional technical proofs for all
lemmas are provided in the supplementary material \cite{fan2014Supp}.

\subsection{Lemmas} \label{sec: SecCovZ}

We introduce the following lemmas which are used in the proofs of
Theorems \ref{Thm2}--\ref{thmm: MR}.

\begin{lemma}\label{lem4} Under model setting (\ref{model}) and the
conditions in Theorem~\ref{Thm2}, for sufficiently large $n$, with
probability at least $1 - p\exp(-C\tilde{\tau}_{1,p}^2 n^{1 -
2\kappa
})$, it holds that
\[
\max_{1 \leq j \leq p}\bigl|\hat{\sigma}^2_j/\tilde{
\sigma}^2_j - 1\bigr| \leq T_{n,p}/6
\]
for some positive constant $C$, where $T_{n,p}$ is the same as in
Theorem~\ref{Thm2}.
\end{lemma}

\begin{lemma} \label{lem: basic-inequality}
Under Condition \ref{con: mean}, we have
%
\begin{equation}
\label{eq: basic-final} \widetilde{C}n^{-1}\|\bX\bdelta\|_2^2
+\operatorname{pen}(\widehat{\btheta}) \leq\bigl\|n^{-1}\bveps^T\bX
\bigr\|_{\infty}\|\bdelta\|_1 + \operatorname {pen}(\btheta_0),
\end{equation}
where $\widetilde{C}$ is some positive constant depending on the
positive constant $\pi_{\min}$ in Condition \ref{con: mean},
$\bdelta
=\widehat{\btheta}-\btheta_0$ is the estimation error for the
regularized estimator
$\widehat{\btheta}$ defined in \eqref{eq: elastic-obj} and $\bveps
=\by
-E(\by|\bX)$ with $\by=(\Delta_1, \ldots, \Delta_n)^T$.
\end{lemma}

%

\begin{lemma} \label{lem: error-X}
Assume that Condition \ref{con: distribution} holds. If $\log(p)=o(n)$, then with probability
$1-O(p^{-\tilde{c}_1})$, we have $\|n^{-1}\bveps^T\bX\|_{\infty
}\leq
2^{-1}c_0\sqrt{\log(p)/n}$,
where $c_0$ is some positive constant and $\bveps=\by-E(\by|\bX)$ with
$\by=(\Delta_1, \ldots, \Delta_n)^T$.
\end{lemma}



\begin{lemma}\label{lem: con-for-con3}
Assume that Conditions \ref{con: distribution} and \ref{con:
RE-population} hold. 
If 
$5s^{1/2}+ 4\lambda_1^{-1}\times\break \lambda_2\|\btheta_0\|_2=O(n^{{\xi}/2})$ and
$\log(p)=o(n^{1/2-2\xi})$ with constant $0\leq\xi<1/4$, then when $n$
is sufficiently large, with probability at least $1-O(p^{-\tilde
{c}_2})$, where $\tilde c_2$ is some positive constant, it holds that
\[
n^{-1/2}\|\bX\bdelta\|_2\geq(\phi/2)\|\bdelta_S
\|_2
\]
for any $\bdelta\in\mathbb{R}^{\tilde{p}}$ satisfying $\|\bdelta
_{S^c}\|
_1\leq4(s^{1/2}+\lambda_1^{-1}\lambda_2\|\btheta_0\|_2)\|\bdelta
_{S}\|
_2$. 
\end{lemma}

\begin{lemma}\label{lem: bound}
Assume that $\bw=(W_1, \ldots, W_p)^T\in\mathbb{R}^p$ is sub-Gaussian.
Then for any positive constant $c_1$, there exists some positive
constant $C_2$ such that
\[
P \Bigl\{\max_{1\leq j\leq p}|W_j|>C_2\sqrt{
\log(p)} \Bigr\}=O\bigl(p^{-c_1}\bigr).
\]
\end{lemma}

%
%

\subsection{Proof of Theorem \texorpdfstring{\protect\ref{Thm2}}{1}} \label{sec: Proof-Th2}
Since we have the inequality
%
\begin{equation}
\label{bench} |\widehat{D}_j - D_j| \leq|
\widehat{D}_j - \widetilde{D}_j| + |
\widetilde{D}_j - D_j|,
\end{equation}
the key of the proof is to show that with overwhelming probability,
$\widehat{D}_j$ and $\widetilde{D}_j$ are uniformly close as $n \to
\infty$. Then together with Proposition~\ref{Prop1}, we can prove the
desired result in 
Theorem~\ref{Thm2}. The same notation $C$ will be used to denote a
generic constant without loss of generality.

We proceed to prove that $\widehat{D}_j$ and $\widetilde{D}_j$ are
uniformly close. By definitions of $\widehat{D}_j$ and $\widetilde
{D}_j$, along with the fact that $|n_{k}/n| \leq1$ for $k = 1$ and
$2$, we decompose the difference between $\widehat{D}_j$ and
$\widetilde
{D}_j$ as
%
\begin{eqnarray}
\label{decomp4} && \max_{1\leq j\leq p}|\widehat{D}_j -
\widetilde{D}_j|
\nonumber
\\[-8pt]
\\[-8pt]
\nonumber
&&\qquad\leq \max_{1\leq j\leq p}\bigl|\log\hat{\sigma}^2_j
- \log\tilde {\sigma }^2_j\bigr| + \sum
_{k = 1}^2 \max_{1\leq j\leq p}\bigl|\log\bigl[
\bigl(\tilde{\sigma }_j^{(k)}\bigr)^2\bigr] -
\log\bigl[\bigl(\hat{\sigma}_j^{(k)}\bigr)^2
\bigr]\bigr|.
\end{eqnarray}

The following argument is conditioning on the event, denoted by
$\mathcal{E}_1$, such that the results hold in Lemma~\ref{lem4}. 
Then $\hat{\sigma}^2_j$ and $\tilde{\sigma}^2_j$ are uniformly close.
Since $x_n^{-1}\log(1 + x_n) \rightarrow1$ as $x_n \rightarrow0$, it
follows that 
\[
\log\bigl(\hat{\sigma}^2_j/\tilde{\sigma}^2_j
\bigr)/\bigl(\hat{\sigma }^2_j/\tilde {
\sigma}^2_j - 1\bigr) \to1
\]
uniformly for all $j$ as $n \to\infty$. Thus, with a sufficiently
large $n$ uniformly over $j$, we have
%
\begin{eqnarray}
\label{error} && P\Bigl(\max_{1\leq j\leq p}\bigl|\log\hat{
\sigma}^2_j - \log\tilde{\sigma }^2_j\bigr|
> T_{n,p}/3|\mathcal{E}_1\Bigr)
\nonumber
\\[-8pt]
\\[-8pt]
\nonumber
&&\qquad\leq P\Bigl(\max_{1\leq j\leq p}\bigl|\hat{\sigma}^2_j/
\tilde {\sigma }^2_j - 1\bigr|> T_{n,p}/6|
\mathcal{E}_1\Bigr) \leq p\exp\bigl(-C \tilde{\tau}_{1,p}^2
n^{1 - 2\kappa}\bigr).
\end{eqnarray}
By a similar argument, we can derive for $k = 1,2$,
\[
P\Bigl(\max_{1\leq j\leq p}\bigl|\log\bigl[\bigl(\hat{\sigma}_j^{(k)}
\bigr)^2\bigr] - \log \bigl[\bigl(\tilde {\sigma}_j^{(k)}
\bigr)^2\bigr]\bigr| > T_{n,p}/3|\mathcal{E}_1\Bigr)
\leq p\exp\bigl(-C \tilde {\tau}_{1,p}^2 n^{1 - 2\kappa}
\bigr).
\]
In view of (\ref{decomp4}), we get
%
\begin{eqnarray*}
&&P\Bigl(\max_{1\leq j\leq p}|\widehat{D}_j -
\widetilde{D}_j| > T_{n,p}|\mathcal{E}_1\Bigr)
\\
&&\qquad\leq P\Bigl(\max_{1\leq j\leq p}\bigl|\log\hat{\sigma}^2_j
- \log\tilde {\sigma }^2_j\bigr| > T_{n,p}/3|
\mathcal{E}_1\Bigr)
\\
&&\qquad\quad{} + \sum_{k = 1}^2 P\Bigl(\max
_{1\leq j\leq p}\bigl|\log\bigl[\bigl(\hat{\sigma }_j^{(k)}
\bigr)^2\bigr] - \log\bigl[\bigl(\tilde{\sigma}_j^{(k)}
\bigr)^2\bigr]\bigr| > T_{n,p}/3|\mathcal {E}_1\Bigr)
\\
&&\qquad\leq p \exp\bigl(-C \tilde{\tau}_{1,p}^2 n^{1 - 2\kappa}
\bigr).
\end{eqnarray*}
By Lemma~\ref{lem4}, $P(\mathcal{E}_1^c) \leq p \exp(-C \tilde{\tau
}_{1,p}^2 n^{1 - 2\kappa})$. It follows that
%
\begin{eqnarray}
\label{simi} P\Bigl(\max_{1\leq j\leq p}|\widehat{D}_j -
\widetilde{D}_j| > T_{n,p}\Bigr) & \leq &P\Bigl(\max
_{1\leq j\leq p}|\widehat{D}_j - \widetilde{D}_j|
> T_{n,p}|\mathcal{E}_1\Bigr) + P\bigl(
\mathcal{E}_1^c\bigr)
\nonumber
\\[-8pt]
\\[-8pt]
\nonumber
& \leq& p \exp\bigl(-C \tilde{\tau}_{1,p}^2
n^{1 - 2\kappa}\bigr).
\end{eqnarray}

Therefore, for any $p$ satisfying $\log p = O(n^{\gamma})$ with $0 <
\gamma< 1 - 2\kappa$ and $\tilde{\tau}_{1,p}^{-2} = o(n^{1 -
2\kappa-
\gamma})$, we get that for large enough $n$,
\[
P\Bigl(\max_{1\leq j\leq p} |\widehat{D}_j -
\widetilde{D}_j| > T_{n,p}\Bigr) %
\leq\exp\bigl(-C \tilde{\tau}_{1,p}^2 n^{1 - 2\kappa}\bigr).
\]

Since the same conditions hold for the matrices $\bSig_2$ and $\bOmega
_2$, using similar arguments we can prove that the same results hold
for the covariates in $\mathcal{A}_2$ with the test statistics
calculated using the transformed data $\bZ\widehat{\bOmega}_2$.
This completes the proof of Theorem~\ref{Thm2}.

\subsection{Proof of Theorem \texorpdfstring{\protect\ref{thmm: oracle}}{2}} \label
{sec: Proof-Th3}

By Theorem~\ref{Thm2}, it is sufficient to show the second part of
Theorem~\ref{thmm: oracle}. The main idea of the proof is to first
define an event which holds high
probability and then analyze the behavior of the regularized estimator
$\widehat{\btheta}$ conditional on that event.

Define $\bveps=\by-E(\by|\bX)$ with $\by=(\Delta_1, \ldots,
\Delta
_n)^T$. Since 
$\log(p)=o(n^{1/2-2\xi})$, it follows from Condition \ref{con:
distribution} and Lemma~\ref{lem: error-X} that for any $\lambda
_1\geq
c_0\sqrt{\log(p)/n}$,
\[
P\bigl\{\bigl\|n^{-1}\bveps^T\bX\bigr\|_{\infty}>
2^{-1}\lambda_1\bigr\} \leq P \bigl\{\bigl\|n^{-1}
\bveps^T\bX\bigr\|_{\infty}> 2^{-1}c_0\sqrt{
\log (p)/n} \bigr\} =O\bigl(p^{-\tilde{c}_1}\bigr),
\]
where $\tilde{c}_1$ is some positive constant. Meanwhile, from Lemma~\ref{lem: con-for-con3}, under the assumptions that $5s^{1/2}+4\lambda
_1^{-1}\lambda_2\|\btheta_0\|_2=O(n^{\xi/2})$ and $\log
(p)=o(n^{1/2-2\xi
})$, we have
$n^{-1/2}\|\bX\bdelta\|_2\geq(\phi/2)\bdelta_S^T\bdelta_S$ for any
$\bdelta\in\mathbb{R}^{\tilde{p}}$ satisfying $\|\bdelta_{S^c}\|
_1\leq
4(s^{1/2}+\lambda_1^{-1}\lambda_2\|\btheta_0\|_2)\|\bdelta_{S}\|_2$
when $n$ is sufficiently large, with probability at least
$1-O(p^{-\tilde{c}_2})$. Combining these two results we obtain that
with probability at least $1-O(p^{-\tilde{c}_1})-O(p^{-\tilde
{c}_2})=1-O(p^{-c_1})$ with $c_1=\min\{\tilde{c}_1, \tilde{c}_2\}$, it
holds simultaneously that
%
\begin{eqnarray}
 \bigl\|n^{-1}\bveps^T\bX\bigr\|_{\infty}
&\leq&2^{-1}\lambda_1, \label{eq:
err-X}
\\
\label{eq: RE-Sample} n^{-1/2}\|\bX\bdelta\|_2&\geq&(\phi/2)
\bdelta_S^T\bdelta_S,
\end{eqnarray}
for any $\lambda_1\geq c_0\sqrt{\log(p)/n}$ and
$\bdelta\in\mathbb{R}^{\tilde{p}}$ satisfying $\|\bdelta_{S^c}\|
_1\leq
4(s^{1/2}+ \break \lambda_1^{-1}\lambda_2\|\btheta_0\|_2)\|\bdelta_{S}\|_2$
when $n$ is sufficiently large. From now on, we condition on the event
that inequalities \eqref{eq: err-X} and \eqref{eq: RE-Sample} hold.

It follows from Condition \ref{con: mean} and Lemma~\ref{lem:
basic-inequality} that
\[
\widetilde{C}n^{-1}\|\bX\bdelta\|_2^2 +
\lambda_1\|\widehat{\btheta}\|_1+ \lambda_2
\|\widehat{\btheta}\|_2^2 \leq\bigl\|n^{-1}
\bveps^T\bX\bigr\|_{\infty}\|\bdelta\|_1 +
\lambda_1\| \btheta _0\|_1+
\lambda_2\|\btheta_0\|_2^2,
\]
where $\widetilde{C}$ is some positive constant.
Thus, by inequality \eqref{eq: err-X}, we have
\[
\widetilde{C}n^{-1}\|\bX\bdelta\|_2^2 +
\lambda_1\|\widehat{\btheta}\|_1+ \lambda_2
\|\widehat{\btheta}\|_2^2 \leq2^{-1}
\lambda_1\|\bdelta\|_1 + \lambda_1\|
\btheta_0\|_1+ \lambda _2\|
\btheta_0\|_2^2.
\]
Recall that $\bdelta=\widehat{\btheta}-\btheta_0$. Adding $2^{-1}\|
\bdelta\|_1-2\lambda_2\btheta_0^T(\widehat{\btheta}-\btheta_0)$
to both
sides of the above inequality and rearranging terms yield
%
\begin{eqnarray}
\label{eq6} && \widetilde{C}n^{-1}\|\bX\bdelta\|_2^2
+2^{-1}\lambda_1\|\bdelta\|_1+
\lambda_2\|\bdelta\|_2^2
\nonumber
\\[-8pt]
\\[-8pt]
\nonumber
&&\qquad\leq \lambda_1\bigl(\|\btheta_0\|_1-\|
\widehat{\btheta}\|_1+\|\widehat {\btheta}-\btheta_0
\|_1\bigr) -2\lambda_2\btheta_0^T(
\widehat{\btheta}-\btheta_0).
\end{eqnarray}
Note that $ \|\btheta_0\|_1-\|\widehat{\btheta}\|_1+\|\widehat
{\btheta
}-\btheta_0\|_1
=\|\btheta_{0, S}\|_1-\|\widehat{\btheta}_S\|_1+\|\widehat{\btheta
}_S-\btheta_{0, S}\|_1$ since $|\btheta_{0, j}|-|\widehat{\btheta
}_j|+|\widehat{\btheta}_j-\btheta_{0, j}|=0$ for all $j\in S^c$. By the
triangle inequality and the Cauchy--Schwarz inequality, we have
%
\begin{equation}
\label{eq7} \|\btheta_0\|_1-\|\widehat{\btheta}
\|_1+\|\widehat{\btheta }-\btheta_0\|_1
\leq2\|\widehat{\btheta}_S-\btheta_{0, S}\|_1
\leq2s^{1/2}\|\bdelta_{S}\|_2.
\end{equation}
Note that $|\btheta_0^T(\widehat{\btheta}-\btheta_0)|
=|\btheta_{0, S}^T(\widehat{\btheta}_S-\btheta_{0, S})|$. An
application of the Cauchy--Schwarz inequality gives
%
\begin{equation}
\label{eq8} \bigl|\btheta_0^T(\widehat{\btheta}-
\btheta_0)\bigr| \leq\|\btheta_{0, S}\|_2 \|\widehat{
\btheta}_S-\btheta_{0, S}\|_2 =\|
\btheta_{0}\|_2 \|\bdelta_{S}\|_2.
\end{equation}
Combining these three results in \eqref{eq6}--\eqref{eq8} yields
%
\begin{equation}
\label{eq9} \widetilde{C}n^{-1}\|\bX\bdelta\|_2^2
+2^{-1}\lambda_1\|\bdelta\|_1+
\lambda_2\|\bdelta\|_2^2 \leq2\bigl(
\lambda_1s^{1/2}+\lambda_2\|
\btheta_0\|_2\bigr)\|\bdelta_S
\|_2,
\end{equation}
which, together with the fact that $\|\bdelta_{S^c}\|_1\leq\|\bdelta
\|
_1$, implies a basic constraint
\[
\|\bdelta_{S^c}\|_1\leq4\bigl(s^{1/2}+
\lambda_1^{-1}\lambda_2\|\btheta
_0\| _2\bigr)\|\bdelta_S\|_2.
\]
Thus, by inequality \eqref{eq: RE-Sample}, we have
$n^{-1/2}\|\bX\bdelta\|_2\geq(\phi/2)\bdelta_S^T\bdelta_S$.
This, together with~\eqref{eq9}, gives
\[
4^{-1}\widetilde{C}\phi^2 \|\bdelta_S
\|_2^2\leq\widetilde {C}n^{-1}\| \bX\bdelta
\|_2^2 \leq2\bigl(\lambda_1s^{1/2}+
\lambda_2\|\btheta_0\|_2\bigr)\|
\bdelta_S\|_2.
\]
Solving this inequality yields $\|\bdelta_S\|_2
\leq8\widetilde{C}^{-1}\phi^{-2}(\lambda_1s^{1/2}+\lambda_2\|
\btheta
_0\|_2)$. Combining this with \eqref{eq9} entails that
\[
\widetilde{C}n^{-1}\|\bX\bdelta\|_2^2
+2^{-1}\lambda_1\|\bdelta\|_1+
\lambda_2\|\bdelta\|_2^2 \leq16
\widetilde{C}^{-1}\phi^{-2}\bigl(\lambda_1s^{1/2}+
\lambda_2\| \btheta _0\|_2
\bigr)^2
\]
holds with probability at least $1-O(p^{-c_1})$. Thus from the above
inequality we have
\begin{eqnarray*}
 \|\widehat{\btheta}-\btheta_0\|_1&=& \|\bdelta
\|_1 \leq32\widetilde{C}^{-1}\phi^{-2}\bigl(
\lambda_1s^{1/2}+\lambda_2\| \btheta
_0\|_2\bigr)^2/\lambda_1,
\\
 n^{-1/2}\bigl\|\bX(\widehat{\btheta}-\btheta_0)
\bigr\|_2&=&n^{-1/2}\|\bX \bdelta\|_2 \leq4
\widetilde{C}^{-1}\phi^{-1}\bigl(\lambda_1s^{1/2}+
\lambda_2\| \btheta _0\|_2\bigr),
\end{eqnarray*}
hold simultaneously with probability at least $1-O(p^{-c_1})$. This
completes the proof of Theorem~\ref{thmm: oracle}.

\subsection{Proof of Theorem \texorpdfstring{\protect\ref{thmm: MR}}{3}} \label{sec: Proof-Th4}

Recall that ${\mathbf z}=(Z_1, \ldots, Z_p)^T=\Delta{\mathbf
z}^{(1)}+(1-\Delta){\mathbf z}
^{(2)}$ and ${\mathbf x}=(1, Z_1, \ldots, Z_p, Z_1^2, Z_1Z_2, \ldots,
Z_{p-1}Z_p, Z_p^2)^T$. Define an event
\[
\mathcal{E}_2 =\bigl\{\|\widehat{\btheta}-\btheta_0
\|_1\leq32\widetilde {C}^{-1}\phi^{-2}\bigl(
\lambda_1s^{1/2}+\lambda_2\|
\btheta_0\| _2\bigr)^2/\lambda
_1\bigr\},
\]
where positive constant $\widetilde{C}$ is given in Theorem~\ref{thmm:
oracle}. From Theorem~\ref{thmm: oracle}, we have $P(\mathcal
{E}_2^c)\leq O(p^{-c_1})$. By Lemma~\ref{lem: bound}, under Condition
\ref{con: distribution}, there exists a positive constant $C_2$ such that
%
\begin{equation}
\label{eq: Zj-bound} P\Bigl\{\max_{1\leq j\leq p}\bigl|Z_j^{(k)}\bigr|>
C_2\sqrt{\log(p)}\Bigr\}\leq O\bigl(p^{-c_1}\bigr)
\end{equation}
for $k=1, 2$, where $(Z_1^{(k)}, \ldots, Z_p^{(k)})^T={\mathbf z}^{(k)}$.
Define an event
$\mathcal{E}_3 =\{\|{\mathbf z}\|_{\infty}\leq C_2\sqrt{\log(p)}\}$.
Then $P(\mathcal{E}_3^c)\leq O(p^{-c_1})$. An application of the
Bonferroni inequality gives
%
\begin{equation}
\label{eq:E1E2} P\bigl(\mathcal{E}_2^c\cup
\mathcal{E}_3^c\bigr)\leq O\bigl(p^{-c_1}\bigr)+O
\bigl(p^{-c_1}\bigr)=O\bigl(p^{-c_1}\bigr).
\end{equation}

Denote by $\mathcal{C}_1$ the event $\{{\mathbf z} \mbox{ from class
1}\}$.
Note that on the event $\mathcal{E}_3$, we have $\|{\mathbf x}\|
_{\infty}\leq
C_2\log(p)$ where we use a generic constant $C_2$ to simplify notation.
Using the property of conditional probability gives
%
\begin{eqnarray}
\label{eq: Rn21} R_n(2|1) &=&P\bigl({\mathbf x}^T\widehat{
\btheta}\leq0|\mathcal{C}_1\bigr) = P\bigl({\mathbf x}^T
\btheta_0\leq{\mathbf x}^T(\btheta_0-
\widehat {\btheta})|\mathcal {C}_1\bigr)
\nonumber
\\[-8pt]
\\[-8pt]
\nonumber
&\leq& P\bigl({\mathbf x}^T\btheta_0\leq{\mathbf
x}^T(\btheta_0-\widehat {\btheta})|\mathcal
{C}_1, \mathcal{E}_2\cap\mathcal{E}_3\bigr)
+ P\bigl(\mathcal{E}_2^c\cup\mathcal{E}_3^c
\bigr).
\end{eqnarray}
Note that conditioning on the event $\mathcal{E}_2\cap\mathcal{E}_3$,
${\mathbf x}^T(\btheta_0-\widehat{\btheta})$ can be bounded as
\begin{eqnarray*}
\bigl|{\mathbf x}^T(\btheta_0-\widehat{\btheta})\bigr| &\leq&\|{
\mathbf x}\|_{\infty} \|\widehat{\btheta}-\btheta_0
\|_1 \\
&\leq&32C^{-1}C_2\phi^{-2}\log(p)
\bigl(\lambda_1s^{1/2}+\lambda_2\| \btheta
_0\|_2\bigr)^2/\lambda_1.
\end{eqnarray*}
Then
$|{\mathbf x}^T(\btheta_0-\widehat{\btheta})|\leq C_3\Lambda_n$
with positive
constant $C_3=32C^{-1}C_2\phi^{-2}$. Thus we have
\begin{eqnarray*}
&& P\bigl({\mathbf x}^T\btheta_0\leq{\mathbf
x}^T(\btheta_0-\widehat {\btheta})|\mathcal
{C}_1, \mathcal{E}_2\cap\mathcal{E}_3\bigr)
\\
&&\qquad\leq P\bigl({\mathbf x}^T\btheta_0\leq C_3
\Lambda_n|\mathcal{C}_1, \mathcal {E}_2\cap
\mathcal{E}_3\bigr)
\\
&&\qquad=  P\bigl({\mathbf x}^T\btheta_0\leq C_3
\Lambda_n|\mathcal{C}_1, \mathcal{E}_3\bigr)
= \frac{P({\mathbf x}^T\btheta_0\leq C_3\Lambda_n, \mathcal
{E}_3|\mathcal
{C}_1)}{P(\mathcal{E}_3|\mathcal{C}_1)}
\\
&&\qquad\leq \frac{P({\mathbf x}^T\btheta_0\leq C_3\Lambda_n|\mathcal
{C}_1)}{P(\mathcal{E}_3|\mathcal{C}_1)} = \frac{F_1(C_3\Lambda_n)}{P(\mathcal{E}_3|\mathcal{C}_1)},
\end{eqnarray*}
where $F_1(\cdot)$ is the cumulative distribution function of
${\mathbf x}
^T\btheta_0|\mathcal{C}_1$. This inequality, together with \eqref{eq:
Rn21}, entails
\[
R_n(2|1)\leq\frac{F_1(C_3\Lambda_n)}{P(\mathcal{E}_3|\mathcal{C}_1)}+ P\bigl(\mathcal{E}_2^c
\cup\mathcal{E}_3^c\bigr).
\]
By the definition of $F_1(\cdot)$, we have $R(2|1)=F_1(0)$. Thus
%
\begin{eqnarray}
\label{eq:diff-R21} && R_n(2|1)-R(2|1)
\nonumber
\\[-8pt]
\\[-8pt]
\nonumber
&&\qquad\leq \frac{F_1(C_3\Lambda_n)-F_1(0)}{P(\mathcal{E}_3|\mathcal
{C}_1)}+ \biggl[\frac{1}{P(\mathcal{E}_3|\mathcal{C}_1)}-1 \biggr]F_1(0)
+P\bigl(\mathcal{E}_2^c\cup\mathcal{E}_3^c
\bigr).
\end{eqnarray}
From Condition \ref{con: CDF}, we have $0\leq\Lambda_n<\epsilon_0$
when $n$ is sufficiently large and $C_3\Lambda_n=o(1)$. It follows that
$F_1(C_3\Lambda_n)-F_1(0) =F'(\Lambda_n^{\ast})\Lambda_n\leq
C_3r_n\Lambda_n$ where $\Lambda_n^{\ast}$ is between $0$ and
$C_3\Lambda
_n$. In view of \eqref{eq: Zj-bound}, we have
\[
P(\mathcal{E}_3|\mathcal{C}_1)=P \Bigl\{\max
_{1\leq j\leq
p}\bigl|Z_j^{(1)}\bigr|\leq C_2
\sqrt{\log(p)} \Bigr\}=O\bigl(p^{-c_1}\bigr).
\]
Combining this with \eqref{eq:E1E2} and \eqref{eq:diff-R21} entails
\begin{eqnarray*}
R_n(2|1)-R(2|1)& \leq& \frac{C_3r_n\Lambda_n}{1-O(p^{-c_1})}+\biggl\llvert
\frac
{1}{1-O(p^{-c_1})}-1\biggr\rrvert +O\bigl(p^{-c_1}\bigr)+O
\bigl(p^{-c_1}\bigr)
\\
&=& O(r_n\Lambda_n)+O\bigl(p^{-c_1}\bigr).
\end{eqnarray*}
Similarly, we can show that $R_n(1|2)\leq R(1|2)+O(r_n\Lambda
_n)+O(p^{-c_1})$. Combining these two results completes the proof of
Theorem~\ref{thmm: MR}.

\section*{Acknowledgments}
The authors sincerely
thank the Co-Editor, Associate Editor and two referees for their valuable
comments that helped improve the article substantially.

\begin{supplement}
\stitle{Supplement to ``Innovated interaction screening for
high-dimensional nonlinear classification''}
\slink[doi]{10.1214/14-AOS1308SUPP} 
\sdatatype{.pdf}
\sfilename{aos1308\_supp.pdf}
\sdescription{We provide additional lemmas and technical proofs.}
\end{supplement}

%

\printaddresses
\end{document}